\documentclass{article}

     \PassOptionsToPackage{numbers, compress}{natbib}

     \usepackage[preprint]{neurips_2020}

\usepackage{neurips_2020}

\usepackage[utf8]{inputenc} 
\usepackage[T1]{fontenc}    
\usepackage{hyperref}       
\usepackage{url}            
\usepackage{booktabs}       
\usepackage{amsfonts}       
\usepackage{nicefrac}       
\usepackage{microtype}      

\usepackage{paralist}
\usepackage{mathtools}
\usepackage{subcaption}
\usepackage{amssymb}
\usepackage{multirow}
\usepackage{tabularx}
\usepackage{xspace}
\usepackage{xcolor}
\usepackage{cleveref}
\usepackage{graphicx}
\usepackage{float}

\newcommand{\xhdr}[1]{\vspace{2pt}\noindent\textbf{#1}}

\DeclareMathSymbol{@}{\mathord}{letters}{"3B}

\newcommand{\mcmi}{MCMI\xspace}

\title{\mcmi: Multi-Cycle Image Translation with\\Mutual Information Constraints}

\author{%
  Xiang Xu\qquad
  Megha Nawhal\qquad
  Greg Mori \qquad
  Manolis Savva \\
  Simon Fraser University \\
  \{xuxiangx,\enspace mnawhal,\enspace mori, \enspace msavva\}@sfu.ca \\
}

\begin{document}
\maketitle
\begin{abstract}
We present a mutual information-based framework for unsupervised image-to-image translation.
Our \mcmi approach treats single-cycle image translation models as modules that can be used recurrently in a multi-cycle translation setting where the translation process is bounded by mutual information constraints between the input and output images.
The proposed mutual information constraints can improve cross-domain mappings by optimizing out translation functions that fail to satisfy the Markov property during image translations.
We show that models trained with \mcmi produce higher quality images 
and learn more semantically-relevant mappings compared to state-of-the-art image translation methods.
The \mcmi framework can be applied to existing unpaired image-to-image translation models with minimum modifications.
Qualitative experiments and a perceptual study demonstrate the image quality improvements and generality of our approach using several backbone models and a variety of image datasets.
\end{abstract}

\section{Introduction}

Image-to-image (I2I) translation has gained prominence in recent years.
The goal for I2I translation is to map an image from one domain to a corresponding image in another domain.
This is particularly challenging in the unpaired case where no aligned training pairs are available.
Most unpaired I2I translation models~\cite{zhu2017toward,zhu2017unpaired,liu2017unsupervised,lee2018diverse,huang2018multimodal,choi2018stargan,lee2019drit++,choi2019stargan} rely on cycle consistency loss to achieve realistic image generation.
The underlying assumption is that semantic information is preserved across domains when the model is forced to reconstruct the original input from the translated output.
However, for tasks with one-to-many mappings (e.g., photo colorization), the use of cycle consistency forces the model to translate back to only one possible outcome.
This makes image generators susceptible to steganography, where information necessary to reconstruct the input is hidden in small perturbations of the output image~\cite{chu2017cyclegan}.
Consequently, the output image is not constrained to be semantically related to the input in order to satisfy cycle consistency.
This leads to less robust I2I models and inferior mapping functions which fail to preserve important semantic information across domains~\cite{chu2017cyclegan,bashkirova2019adversarial}.

In this work, we present an information-theoretic view of the problem and propose an I2I translation framework to address the issues with cycle consistency loss.
Our insight is that when mapping from source to target domain, an ideal I2I model should only rely on semantic information from the source and not from any other extra data. We reformulate an I2I model that satisfy this constraint as a Markov chain and demonstrate that cycle consistency loss violates the Markov property as output and input images are forced to reconstruct each other.
Ideally, we would like the image quality benefits from cycle consistency while also enforcing the Markov property to preserve semantic information.

To reconcile the Markov property and cycle consistency, we extend single-cycle translation to multi-cycle translation, proposing multi-cycle translation with mutual information constraints (\mcmi).
Specifically, we propose multi-cycle translation where the first translation cycle employs cycle consistency loss and we use the \emph{data processing inequality theorem} %
to enforce non-increasing mutual information (MI) for subsequent cycle translations.
Any learned mapping function that does not satisfy this inequality will not correspond to a Markov chain and thus will fail to satisfy the Markov property. %
This avoids the function space where the Markov property will not be satisfied during training. %
Together with the cycle consistency loss in the first cycle, our \mcmi framework results in higher quality image translation and semantically-relevant cross-domain mapping functions while adhering to the Markov property.
To summarize, our contributions are as follows.
\begin{compactitem}
\item We propose an information-theoretic formulation of unpaired image-to-image translation: the multi-cycle mutual information (\mcmi) framework.
\item We demonstrate the generality of \mcmi by applying it to various single-cycle I2I models.
\item We perform quantitative and qualitative experiments to show the \mcmi framework produces higher quality images and more semantically-related cross-domain mappings compared to the backbone single-cycle I2I models.
\end{compactitem}

\section{Related Work}

Image-to-image (I2I) translation maps an image from a source domain to an image in a target domain.
Paired I2I translation refers to the setting where a dataset contains input-output pairs for learning the translation function.
Prior work in this setting has used non-parametric models~\cite{efros1999texture}, parametric models such as CNNs\cite{sangkloy2017scribbler}, and conditional GANs\cite{pix2pix2016,wang2019learning,zhai2019lifelong}.
Unpaired I2I translation refers to the problem setting where input-output pairs are not available.
This setting implicitly requires learning relations between the source and target domains.
Prior work tackling this problem spans a diverse array of techniques such as patch-based Markov Random Fields~\cite{rosales2003unsupervised}, cross-modal weight sharing networks~\cite{liu2016coupled,aytar2017cross}, and combination of VAEs and GANs~\cite{liu2017unsupervised,taigman2016unsupervised,shrivastava2017learning,bousmalis2017unsupervised}.
All these approaches focus on one-to-one mapping between the source and target domain.

Recently, there has been some work on one-to-many image translation~\cite{huang2018multimodal,lee2018diverse,lee2019drit++,zhu2017multimodal}.
MUNIT~\cite{huang2018multimodal} assumes content features share similar values across different domains;
DRIT~\cite{lee2018diverse} further use a content discriminator to force the content features to be in the same latent space; DRIT++~\cite{lee2019drit++} extends DRIT with mode-seeking loss~\cite{mao2019mode}; and
DMIT~\cite{yu2019multi} provides an efficient model for both multi-modal and multi-domain translation. 

Cycle consistency loss is widely used in the aforementioned unpaired I2I models.
The issues with cycle consistency loss are first explored by \citet{chu2017cyclegan} demonstrating that CycleGAN might lead to steganography using perturbations in the output to hide information on the input.
Later work by \citet{bashkirova2019adversarial} formulate the problem as a self-adversarial attack. However, their method is inefficient as the model needs further training on adversarial images with injected noise.

\section{MCMI: Multi-Cycle Translation with Mutual Information Constraints}

Our goal is to perform image-to-image (I2I) translation in an unpaired setting where the input-output pairings are not available.
Given an image $x$ from domain $X$, we produce a set of generated images $\mathcal{Y}$ for a domain $Y$.
For example, $X$ could be the domain of \emph{cat} images, while $Y$ could be the domain of \emph{dog} images.
Given a particular instance of a cat image $x$, there are many corresponding dog images $y \in \mathcal{Y}$ that are potential translations (i.e. one-to-many mapping).
Our approach focuses on ensuring two key properties in the output set $\mathcal{Y}$: i) it contains diverse elements that are realistic images of the domain $Y$; and ii) it preserves semantically-relevant content (e.g., animal pose or expression) during translation while allowing variation in visual content (e.g., breed of dog or fur color).

To learn semantically-relevant cross-domain mappings, the mapping of visual features should only share dependencies between the input and generated images within every cycle, but not with the sequence of generated data that preceded it from earlier cycles.
Therefore, we formulate this problem as a Markov chain where the Markov property is satisfied across domain translations.
Direct optimization of the Markov property is difficult.
Such optimization would require enforcing conditional independence of future generated images from previously generated ones, given the current generated image. However, the inputs and outputs are conditionally dependent in an unconstrained I2I translation network.
Instead, we propose the multi-cycle MI constraints that are easier to optimize.

Our framework is based on a multi-cycle Markov chain formulation of the I2I translation problem which we describe first in \Cref{sec:markov_chain}.
Then, we describe the components comprising the \mcmi framework: i) a backbone single-cycle I2I module (\Cref{sec:i2i_translation}); ii) an MI estimator (\Cref{sec:mi_estimation}); iii) MI constraints (\Cref{sec:mi_constraints}); and iv) an I2I loss function forcing the MI constraints (\Cref{sec:mi_loss}).

\subsection{Multi-Cycle Image-to-image Translation as a Markov Chain}
\label{sec:markov_chain}

We first extend single-cycle translation to multi-cycle translation by treating an I2I translation model as a backbone module.
Every translation cycle consists of $X^n \rightarrow Y^n \rightarrow X^{n+1}$.
Here $X$, $Y$ refer to two image domains, and the superscript indicates how many times a particular domain has been visited.
At each cycle translation step, $X^{n+1}$ is referred to as the reconstructed image of input $X^n$ and $Y^n$ is the translated image.
The reverse translation going from $Y$ to $X$ is similar.
For simplicity, we do not explicitly state the reverse process.

We propose to view an optimal multi-cycle I2I translation model as a Markov chain.
This means that the reconstructed image $X^{n+1}$ should only depend on the translated image $Y^n$ and is independent of input $X^n$ and all other images before it. Here, we treat image domains as random variables and visits to the same domain at different cycles as different random variables.
This Markov chain assumption addresses the issues introduced by the cycle consistency loss as future image generations no longer rely on encoded information from the initial input. Under this assumption, the I2I model avoids encoding information from earlier translations and learning of trivial mapping functions.

However, the commonly-used cycle consistency loss violates the Markov property as it directly optimizes the reconstructed image to be close to the input with a direct pixel-level loss. %
Simply eliminating the cycle consistency loss causes decreased image quality~\cite{bashkirova2019adversarial}.
To reconcile the Markov property and cycle consistency, 
we formulate our loss functions within a two-cycle translation: $X^1 \rightarrow Y^1 \rightarrow X^2 \rightarrow Y^2 \rightarrow X^3$.
For the first cycle, we apply the cycle consistency loss (encouraging $X^2$ close to $X^1$).
For the second cycle, we only use the MI constraints between output $Y^2, X^2, X^3$ and input $X^1$ to characterize and bound the cross-domain mappings and achieve the desired Markov property.
The same I2I module is used in both cycles so there is no increase in network parameters.
We empirically find that two cycles offer a good trade-off between performance, memory use, and efficiency. Please refer to the supplementary material for more details.

Directly optimizing towards a Markov chain is difficult.
We instead use the \emph{data processing inequality theorem}~\cite{beaudry2011dpi} which states that given a Markov chain, the MI must be non-increasing going from input to the output. \mcmi enforces the non-increasing MI constraints in the second cycle with $I(X^1;X^2) \geq I(X^1;Y^2)\geq I(X^1;X^3)$.
Any learned mapping function that does not satisfy this inequality will not correspond to a Markov chain and fails to satisfy the Markov property.
Explicitly optimizing for non-increasing MI will train a model that avoids the function space where the Markov property is not satisfied.
Combined with the cycle consistency loss in the first cycle, we obtain better quality image translations and semantically-relevant cross-domain mapping functions.%

\subsection{Backbone Module}
\label{sec:i2i_translation}

\mcmi starts with a backbone I2I module %
that can be any of the existing models in the literature.
We examine three state-of-the-art models: CycleGAN~\cite{zhu2017unpaired}, MUNIT~\cite{huang2018multimodal} and DRIT++~\cite{lee2019drit++}.
Here, we describe the key differences between these modules.

\xhdr{CycleGAN} consists of two image mapping functions $G: X \rightarrow Y$ and $F: Y \rightarrow X$.
The self-reconstruction loss is used together with two image domain discriminators to train the model.
The mapping functions $G$ and $F$ are deterministic and do not lead to one-to-many mapping. Refer to \citet{zhu2017unpaired} for more details.

\xhdr{MUNIT} makes the assumption that image features can be disentangled into two latent spaces: a domain-invariant content space, and a domain-specific style space.
During training, %
only the encoded content feature is shared across domain.
The style feature is swapped at each step.
Cycle consistency is enforced after one full cycle translation.
At test time, style feature can be randomly sampled from $\mathcal{N}(0, I)$ and this allows one-to-many mappings.
For more details refer to \citet{huang2018multimodal}.

\xhdr{DRIT++} is based on DRIT~\cite{lee2018diverse} with feature disentanglement and cross-domain mapping similar to MUNIT.
A content discriminator is introduced to ensure the content feature from different domains map to the same content latent space.
A mode seeking loss~\cite{mao2019mode} %
also encourages the generators to explore more modes during training and leads to more diverse one-to-many image mappings based on style features.
Refer to \citet{lee2019drit++} for more details.

\subsection{Mutual Information Estimator}
\label{sec:mi_estimation}

The mutual information $I(X; Y)$ is equivalent to the Kullback-Leibler (KL) divergence between the joint distribution $\mathbb{P}_{X Y}$ and the product of the marginals $\mathbb{P}_{X}\otimes \mathbb{P}_{Y}$. It is defined as: $I(X; Y) =  D_{KL}(\mathbb{P}_{X Y} || \mathbb{P}_{X}\otimes \mathbb{P}_{Y})$.
In practice, computing the exact MI between two high-dimensional random variables (e.g., two images) is intractable. Therefore, approximate algorithms have been used to compute a lower bound of the MI.
MINE~\cite{belghazi2018mine} uses the Donsker-Varadhan representation of KL divergence to derive a lower bound of the MI: $I_{\theta}(X;Y) \leq I(X;Y)$ expressed as $I_{\theta}(X; Y) =  \sup \enskip (\mathbb{E}_{\mathbb{P}_{X Y}}[T_\theta]) - \log(\mathbb{E}_{ \mathbb{P}_{X}\otimes \mathbb{P}_{Y}}[e^{T_\theta}])$. InfoNCE~\cite{oord2018representation} is another approach for estimating the MI.
With $K$ samples of $(x, y)$ from two different random variables $X$, $Y$, the lower bound is represented as:
\begin{equation} \label{eq:mi_infonce}
\begin{split}
I(X; Y) & \geq \mathbb{E}\left[ \frac{1}{K} \sum_{i=1}^{K} \text{log} \frac{e^{f(x_i,y_i)}}{\frac{1}{K} \sum_{j=1}^{K} e^{f(x_j,y_i)}}\right] \triangleq I_{\text{NCE}}^{\text{lower}},
\end{split}
\end{equation}
where $f(x,y)$ is a critic function approximating the density ratio that preserves the MI between $X$, $Y$~\cite{oord2018representation}.
Alternatively, \citet{poole2019variational} derive an upper bound on the MI expressed as:
\begin{equation}\label{eq:mi_upper_infonce}
\begin{split}
I(X; Y) & \leq \mathbb{E}\left[ \frac{1}{K} \sum_{i=1}^{K} \text{log} \frac{p(y_i | x_i)}{\frac{1}{K-1} \sum_{j \neq i} p(y_i | x_j)}\right],
\end{split}
\end{equation}
where $K$ represents the number of samples and $p(y | x)$ is the conditional probability.
Given that an optimal critic for $I_{\text{NCE}}^{\text{lower}}$ is achieved when $f(x,y) = \text{log}\ p(y|x)$~\cite{oord2018representation,poole2019variational}, \Cref{eq:mi_upper_infonce} with near optimal critic can be re-written as:
\begin{equation} \label{eq:mi_upper_infonce_sub}
\begin{split}
I(X; Y) & \leq \mathbb{E}\left[ \frac{1}{K} \sum_{i=1}^{K} \text{log} \frac{e^{f(x_i, y_i)}}{\frac{1}{K-1} \sum_{j \neq i} e^{f(x_j,y_i)}}\right] \triangleq I_{\text{NCE}}^{\text{upper}}.
\end{split}
\end{equation}

The key difference between $I_{\text{NCE}}^{\text{lower}}$ and $I_{\text{NCE}}^{\text{upper}}$ is whether or not $f(x_i, y_i)$ is in the denominator.
This means that once we have a good estimate of the lower bound with a near optimal critic, the upper bound is also known and the actual MI is bounded between these two values. 

We use \Cref{eq:mi_infonce,eq:mi_upper_infonce_sub} based on InfoNCE to estimate the lower and upper bounds of the MI. In our setting, $(x, y)$ are images sampled from different translation steps.
In \Cref{sec:mi_constraints} we will use InfoNCE as a tight constraint to support our multi-cycle MI constraints.
We decided against MINE as it only gives a lower bound on the MI, resulting in worse performance.
The supplemental material provides an ablation experiment to study the impact of each of the bounds.

We define the critic function $f(x,y)$ as the cosine similarity between two image representations after passing through a shared convolutional network $\text{conv}()$. Concretely, $f(x,y) = |\cos(\text{conv}(x), \text{conv}(y))| \cdot s - m,$
where $s$ is a scale and $m$ is an offset.
We use the absolute cosine similarity because two image features encoded by opposite vectors (i.e. cosine similarity of $-1$) should have high MI as knowing one allow us to know the other with high confidence.
The smallest MI value occurs when cosine similarity equals zero, indicating orthogonality or no correlation.
The scale and offset are used to ensure the largest conditional probability $p(y|x)$ is at most $1$.
We set both $s$ and $m$ to $2.0$ in our experiments by hyper-parameter tuning on the validation set.

\Cref{eq:mi_upper_infonce_sub} is most accurate in the optimal critic case, thus during training we have a additional step that maximizes the lower bound $I^\text{lower}_{\text{NCE}}$ to ensure that the critic stays close to optimal.
In summary, we can use these two bounds to estimate MI across translation steps, which can in turn be used in training our I2I model. Detailed training pipeline is provided in the supplementary material.

\subsection{Multi-Cycle Mutual Information Constraints}
\label{sec:mi_constraints}

We use the estimated MI (using neural estimation) to derive an optimization objective for non-increasing MI.
We enforce the MI between generated images at the current translation step and the initial ground truth images to be non-increasing. %
Formally, MI at translation step $n$: $X^n \rightarrow Y^n \rightarrow X^{n+1}$, can be written as $I(X^1; X^{n}) \geq I(X^1; Y^{n})$ and $I(X^1; Y^{n}) \geq I(X^1; X^{n+1}).$
In the previous section we discussed how MI cannot be accurately estimated and hence is difficult to use directly in a loss function.
Thus, we use our estimated MI lower and upper bounds.
To show that the new inequality constraints still achieve non-increasing MI, we note the following holds for any MI lower and upper bounds: $I^\text{lower}_{(q;p)} \geq I^\text{upper}_{(m;n)} \implies I(q;p) \geq I(m;n).$
That is, if the lower bound MI for $(q,p)$ is larger or equal to the upper bound MI for $(m,n)$, then the actual MI for $(q,p)$ is guaranteed to be larger or equal to the actual MI for $(m,n)$.
This means that we can strictly optimize towards non-increasing MI constraints with the lower and upper bounds from \Cref{eq:mi_infonce,eq:mi_upper_infonce_sub}.
In our experiments, we found that enforcing the MI constraints only on the second cycle is sufficient to achieve good image quality.
We can now reformulate the MI constraints for $X^1 \rightarrow Y^1 \rightarrow X^2 \rightarrow Y^2 \rightarrow X^3$ as:
\begin{equation} \label{eq:cycle_mi_one_step_lu}
\begin{split}
I^\text{lower}_{\text{NCE}}(X^1;X^2) & \geq I^\text{upper}_{\text{NCE}}(X^1;Y^2)\\
I^\text{lower}_{\text{NCE}}(X^1;Y^2) & \geq I^\text{upper}_{\text{NCE}}(X^1;X^3).
\end{split}
\end{equation}

\subsection{Overall \mcmi Loss Function}
\label{sec:mi_loss}

The MI loss function for a two-cycle translation model is defined as:
\[
L_{\text{\tiny MI}} = \max\left( I^\text{upper}_{\text{NCE}}(X^1;Y^2) - I^\text{lower}_{\text{NCE}}(X^1;X^2), 0 \right) + \max\left( I^\text{upper}_{\text{NCE}}(X^1;X^3) - I^\text{lower}_{\text{NCE}}(X^1;Y^2), 0 \right),
\]
with the other translation direction swapping the order of images from the two domains.
In addition to $L_{\text{\tiny MI}}$, we use an image adversarial loss $L_\text{\tiny adv}$ to promote realism of the generated image in the second cycle.
We also increase the lower bound of our estimated MI.
This ensures the generated images are bounded and that we are training towards the optimal critic for $I_{\text{NCE}}$.
The overall loss function for \mcmi is:
\begin{equation}\label{eq:overall_loss}
L_{\text{\tiny MCMI}} = L_\text{orig} + L_\text{adv} + \alpha L_{\text{\tiny MI}} - \beta I_\text{lower}.
\end{equation}
Here, $L_\text{\tiny orig}$ refers to the original loss function used in an I2I translation model backbone (CycleGAN, MUNIT and DRIT++).
This includes the image adversarial loss, the content adversarial loss in DRIT++, and the cycle consistency loss applied to all first cycle translations.
Our approach does not rely on a specific I2I translation model, as we do not make any modification to the networks or loss functions within the I2I translation module.
$\alpha$ and $\beta$ are hyper-parameters that control the trade-off between different loss terms. 

\section{Experiments}
\label{sec:experiments}

We demonstrate the effectiveness of the \mcmi framework through quantitative and qualitative evaluation.
We show that our approach can synthesize more realistic and semantically-related images for a diverse set of unpaired I2I translation models and datasets.
Ablation experiments analyze the impact of the multi-cycle MI constraints versus cycle consistency loss.

\subsection{Experimental Procedure} 

\xhdr{Evaluation metrics.}
We define three metrics to evaluate \mcmi relative to baseline I2I models:
\begin{compactitem}
    \item \textbf{Quality.} We measure image quality and realism in two ways. Firstly, we use the Fr{\'e}chet Inception Distance (FID)~\cite{heusel2017gans} which has been shown to correlate with image quality. We compute FID scores from samples generated at the first (FID1) and second (FID2) cycles. Secondly, we perform a user study that involves a `quality' preference judgment. 
    \item \textbf{Diversity.} We use LPIPS~\cite{zhang2018unreasonable} to measure realism and diversity of the generated images. We follow the evaluation setup used for StarGAN v2~\cite{choi2019stargan} where $10$ images are randomly generated per test image and the average of the pairwise distances are computed among all outputs generated from the same input. We do not find significant differences in LPIPS for samples in the first and second cycles and report the LPIPS on first cycle samples. CycleGAN models do not support one-to-many translation so we compute LPIPS from $4$ samples generated at different cycles of the translation (first to fourth cycles).
    \item \textbf{Markovness.} On datasets with ground truth image pairs, we compute the pixel-wise mean squared error $\epsilon_Y$ for the translated image in the target domain and $\epsilon_X$ for the reconstructed image in the source domain. Note that images are normalized to $[0,1]$. We define the ratio $\epsilon_{\text{\tiny mark}} = \epsilon_Y / \epsilon_X$ as a `Markovness error'. High $\epsilon_{\text{\tiny mark}}$ indicates the model reconstructs input $X$ while not producing translated images that match $Y$ well, indicating higher deviation from the Markov assumption. %
\end{compactitem}

\begin{figure*}[t]
\includegraphics[width=\textwidth]{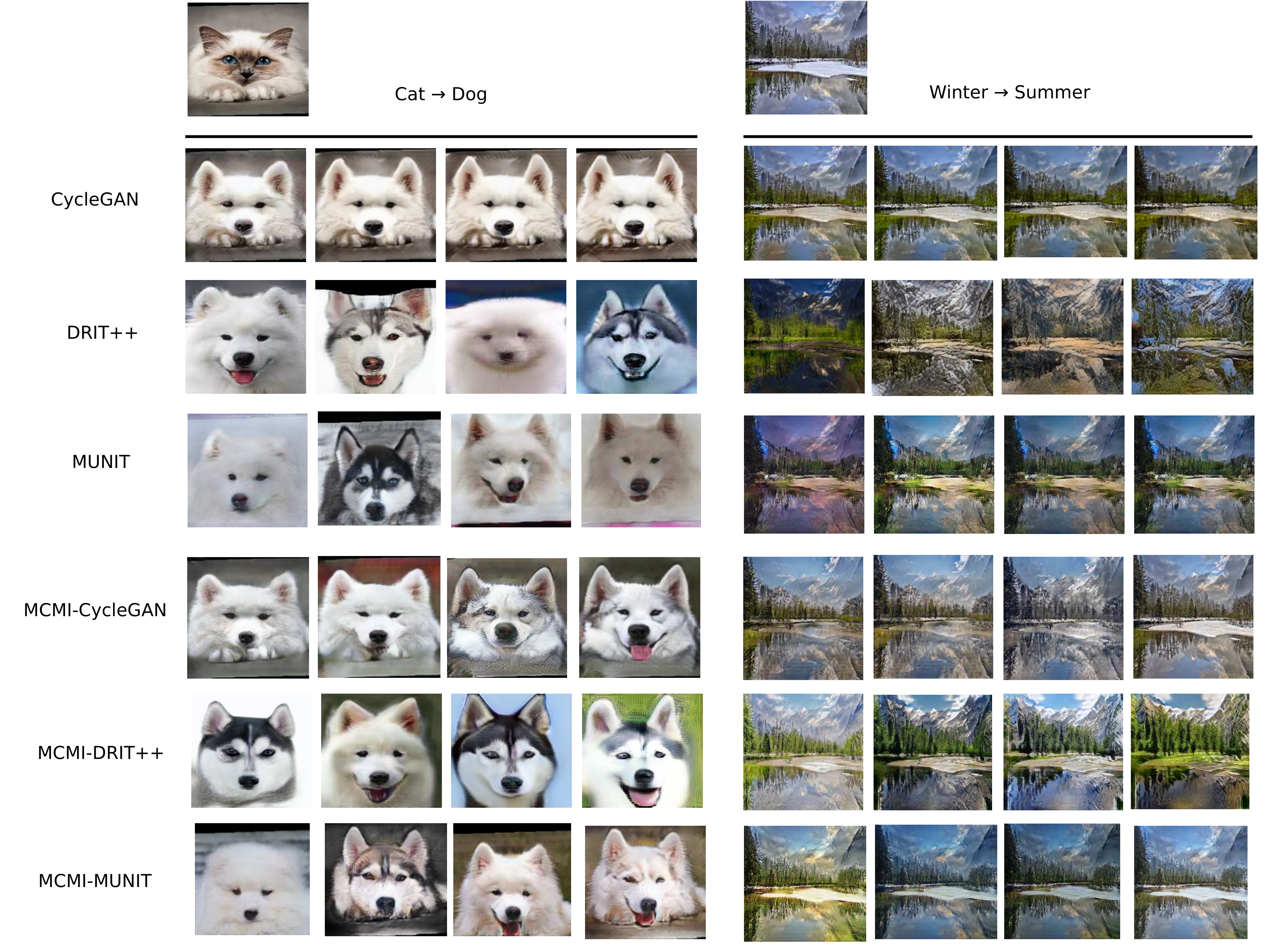}
\caption{
Images generated with \mcmi (two-cycle) and corresponding single-cycle baselines.
The top-left image is the input image.
Samples for CycleGAN models are from successive output images after multiple translation steps.
Samples for MUNIT and DRIT++ models come from different style codes at the first cycle.
The \mcmi-adapted models produce higher quality, diverse images.
Additional examples in the supplement.
}
\label{fig:qualitative}
\end{figure*}

\xhdr{Datasets.}
We evaluate on commonly used I2I datasets including the pets dataset (Cat$\rightarrow$Dog)~\cite{zhu2017unpaired}, Yosemite dataset (Winter$\rightarrow$Summer scenes)~\cite{zhu2017unpaired}, CelebA~\cite{liu2015deep}, and the Google Aerial Photo to Maps dataset~\cite{zhu2017unpaired}.
These datasets span different scales and have diverse attributes and structural correspondences.

\xhdr{Implementation details.}
We implement our proposed method in PyTorch~\cite{paszke2017automatic}.
We use the official PyTorch implementations of the CycleGAN, MUNIT, DRIT++, and StarGAN models.
Images are randomly cropped to $256$x$256$ for CycleGAN and MUNIT, $216$x$216$ for DRIT++ and $128$x$128$ for StarGAN.
The feature encoder $f(x,y)$ used to compute the MI estimates uses four convolutional layers with instance batch normalization and leaky ReLU activation.
Optimizer and I2I network parameters are the same as the corresponding original implementation.

\xhdr{Baselines.}
We compare \mcmi-adapted I2I translation modules against baselines using each module in isolation.
We adapt each baseline to the \mcmi setup according to \Cref{eq:overall_loss}.
CycleGAN~\cite{zhu2017unpaired}, MUNIT~\cite{huang2018multimodal} and DRIT++~\cite{lee2019drit++} are the baselines using a standard single-cycle setup.
MCMI-CycleGAN, MCMI-MUNIT, and MCMI-DRIT++ are the corresponding \mcmi-adapted models.

\begin{table*}[t]
\renewcommand{\arraystretch}{1.3}
\centering
\caption{Comparison of \mcmi-adapted CycleGAN, MUNIT and DRIT++ models against baselines. Arrows indicate whether lower ($\downarrow$) or higher ($\uparrow$) value is better. Standard deviations on two trained models and 4 random seeds indicated for FID. All standard deviations for LPIPS are $\leq 0.003$. We see that \mcmi achieves better image quality (measured by FID), while preserving diversity (LPIPS).}
\vspace{2pt}
\resizebox{\linewidth}{!}{
\begin{tabular}{@{} l @{\hspace{20pt}} ccc ccc @{}}
\toprule
 & \multicolumn{3}{c}{Cat$\rightarrow$Dog} & \multicolumn{3}{c}{Winter$\rightarrow$Summer}\\
 \cmidrule(l){2-4} \cmidrule(l){5-7}
Model & FID1$\downarrow$  & FID2$\downarrow$ & LPIPS$\uparrow$~ & FID1$\downarrow$ & FID2$\downarrow$& LPIPS$\uparrow$~\\\midrule
CycleGAN    & $78.1\pm0.57$ &$70.9\pm0.61$ & $0.09 $ & $65.0\pm0.66$ & $61.5\pm0.75$ & $0.07$ \\
MUNIT       & $21.4\pm1.05$ &$25.8\pm1.04$ & $0.31$ & $55.5\pm0.04$ &$63.4\pm0.05$& $0.22$\\
DRIT++      & $21.9\pm0.41$ &$22.1\pm0.35$ & $\mathbf{0.32}$ &$39.0\pm0.11$ &$44.0\pm0.09$  & $0.24$\\\midrule
\mcmi-CycleGAN & $62.9\pm0.34$ &$28.0\pm0.34$ & $0.22$ & $63.2\pm0.58$ & $60.4\pm0.45$ & $0.18$ \\
\mcmi-MUNIT    & $18.6\pm0.21$ &$21.9\pm0.20$ & $0.30$ &  $48.3\pm0.22$ & $50.2\pm0.21$ & $0.24$ \\
\mcmi-DRIT++   & $\mathbf{17.3\pm0.25}$ &$\mathbf{17.2\pm0.27}$ & $\mathbf{0.32}$ &$\mathbf{37.2\pm0.13}$ & $\mathbf{41.5\pm0.17}$ & $\mathbf{0.25}$\\
\bottomrule
\end{tabular}
}
\label{tab:combined}
\end{table*}

\subsection{Results}

We discuss our extensive qualitative and quantitative evaluation here. Please refer to the supplemental material for additional results.

\xhdr{Qualitative comparisons.}
\Cref{fig:qualitative} shows generated images from baseline single-cycle I2I models and two-cycle \mcmi-adapted counterparts.
Images for CycleGAN models are sampled from different cycles (as explained in the definition of our diversity metric).
Images for DRIT++ and MUNIT-based models come from the first cycle with different sampling of the style code.
We observe that \mcmi leads to higher image quality while preserving the structural content of the input (e.g., head pose for pets and slowly changing landscape for Yosemite images).
For instance, the dog images from the MUNIT baseline have different head orientations.
In contrast, \mcmi-MUNIT maintains the head pose of the input.
We also observe that \mcmi-CycleGAN generates diverse images from different cycles despite the backbone model only supporting one-to-one mapping.
For example, the dog fur color changes and the dog mouth opens up through the translation cycles.

\xhdr{Quantitative comparisons.}
\Cref{tab:combined} reports the quantitative evaluation of \mcmi-adapted models against baseline I2I models. Here, we use all test set images for evaluation.
We see that \mcmi improves image quality (lower FID scores) while retaining diversity (similar LPIPS scores).  
The improvement is particularly pronounced for the Cat$\rightarrow$Dog dataset where there are many high-level visual feature changes during translation.
We also see that DRIT++ and MUNIT models have similar FID in the first and second cycles, whereas CycleGAN has much better FID in the second cycle where the translation is only supervised with our proposed \mcmi constraints.
We believe this is caused by CycleGAN I2I backbone module not supporting one-to-many translation.

\xhdr{User study.}
We generated sets of $4$ images using the \mcmi-DRIT++ model and corresponding DRIT++ baseline for $20$ randomly sampled images cat images in the cat$\rightarrow$dog dataset.
The images in each set are the generation outputs in the target domain after each translation step (from $1$ step to $4$ steps).
We then recruited $10$ participants for our study.
For each such test instance of a source image and two target generated sets (one produced by \mcmi-DRIT++ and one by DRIT++), the participants were asked to select which of the two sets has: i) higher quality, ii) higher overall set diversity. %
The \mcmi-DRIT++ image sets were chosen as having higher quality $79\%$ on average (remaining $16\%$ for baseline DRIT++ images, and $5\%$ ambiguous), while they were chosen to be more diverse $46\%$ of the time ($52\%$ for DRIT++, and $2\%$ ambiguous).
These results indicate that \mcmi generates more realistic images at comparable overall diversity.%

\begin{figure}[H]
\includegraphics[width=1.0\textwidth]{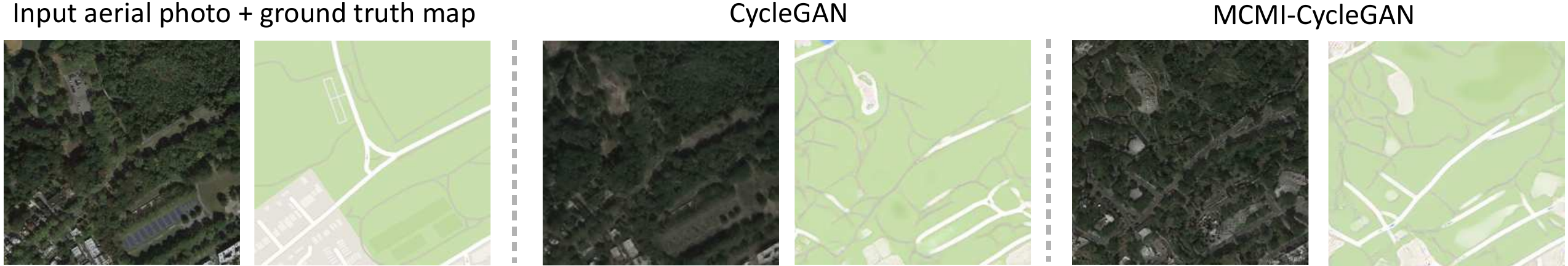}
\caption{Aerial photo to map translations. \mcmi-CycleGAN better preserves semantic information from generated maps to reconstructed photo, exhibiting more Markovian behavior.
}
\label{fig:maps}
\end{figure}

\begin{table*}
\centering
\caption{Average mean squared error for translated maps ($\epsilon_Y$) and reconstructed aerial photos ($\epsilon_X$) at first and second cycles. The Markovness error $\epsilon_{\text{\tiny mark}}$ is defined as $\epsilon_Y / \epsilon_X$ for each cycle.}
\begin{tabular}{@{}l llllll@{}}
\toprule
& \multicolumn{3}{c}{First cycle} & \multicolumn{3}{c}{Second cycle}\\
\cmidrule(lr){2-4} \cmidrule(lr){5-7}
& $\epsilon_Y\downarrow$ & $\epsilon_X$ & $\epsilon_{\text{\tiny mark}}\downarrow$ & $\epsilon_Y\downarrow$ & $\epsilon_X$ & $\epsilon_{\text{\tiny mark}}\downarrow$ \\
\midrule
CycleGAN & $0.0238$ & $0.0056$ & $4.25$ & $0.0243$ & $0.0076$ & $3.20$\\
\mcmi-CycleGAN & $\mathbf{0.0211}$ & $0.0141$ & $\mathbf{1.50}$ & $\mathbf{0.0201}$ & $0.0169$ & $\mathbf{1.19}$\\
\bottomrule
\end{tabular}
\label{tab:markov-chain}
\end{table*}

\xhdr{Markovness comparison.}
\label{sec:markovness}
To measure the degree to which \mcmi helps to enforce the Markov property, we compare the CycleGAN baseline with \mcmi-CycleGAN on the Google Aerial Photo to Maps dataset where paired images are available.
During training, we still train with randomly shuffled unpaired images, but at evaluation time we can compute $\epsilon_\text{\tiny mark}$ using the ground truth pairs. \Cref{fig:maps} provides qualitative examples.
CycleGAN produces map translations that differ significantly from the ground truth while the reconstructed aerial photos are similar to the input.
This indicates that CycleGAN is far from Markovian.
On the other hand, \mcmi-CycleGAN more faithfully utilizes semantic information from the previous input.
For instance, the lower left corner of the \mcmi-CycleGAN translated map has three side roads that connect to the long main road rather than two.
The reconstructed aerial photo also reflects this change.
Quantitative evaluation in \Cref{tab:markov-chain} confirms that \mcmi-CycleGAN is more Markovian. On all test images, \mcmi-CycleGAN achieves both the lowest pixel-wise MSEs for translated maps and Markovness errors at first and second cycle. %

\begin{figure}
\includegraphics[width=0.94\textwidth]{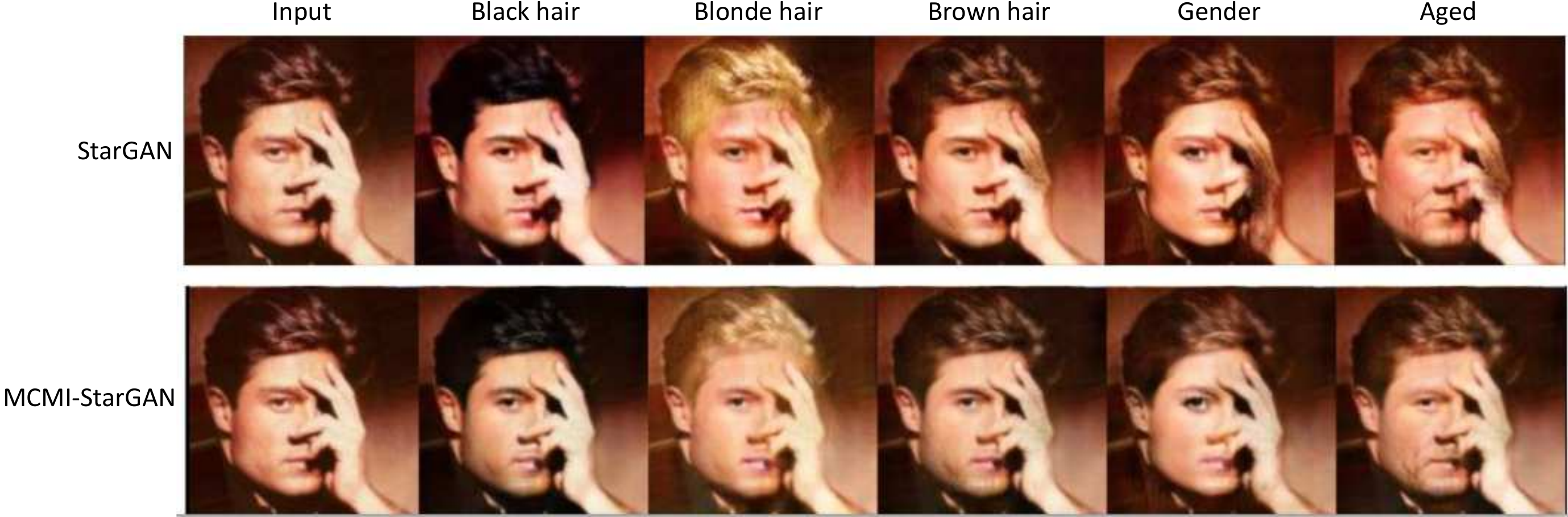}
\caption{Multi-domain I2I translation on the CelebA~\cite{liu2015deep} dataset.
Comparisons between generated images from our \mcmi-adapted StarGAN and the baseline StarGAN model.
}
\label{fig:stargan}
\end{figure}

\xhdr{Multi-domain I2I translation.}
To further demonstrate the generality of our approach we also train a \mcmi-StarGAN model on a multi-domain I2I translation task using the CelebA datasets.
The CelebA~\cite{liu2015deep} dataset is a face attributes dataset with more than $200$K celebrity images, each with $40$ attribute annotations.
We follow the setup for StarGAN~\cite{choi2018stargan} and train on five selected attributes.
\Cref{fig:stargan} shows the results.
Overall, \mcmi-StarGAN produces better quality images.
For example, we see that baseline model generates a lot of hair in regions where should not be (third column), and in cases where part of the face is occluded (e.g., by a hand), StarGAN generates noisy images around the occluded region while \mcmi-StarGAN does not generate black marks.

\section{Conclusion}

We present a mutual information-based approach for the unsupervised image-to-image (I2I) translation problem.
Our \mcmi framework extends several state-of-the-art I2I modules to enable multi-cycle I2I translation bounded by mutual information constraints.
Qualitative and quantitative experiments show that the generated images exhibit increased quality at similar overall diversity, and a human user study further confirms the quality improvements.
We show that \mcmi-adapted I2I models are more Markovian in that they better utilize semantic information from input images during cross-domain mapping.
Though we have focused on unsupervised I2I translation, our approach can be extended to estimate and bound mutual information changes when additional supervision is available in the form of image labels and annotated attributes.
We believe our information theoretic view can be applied to problem domains other than images.


{\small
\bibliographystyle{abbrvnat}
\bibliography{egbib}

\begin{thebibliography}{33}
\providecommand{\natexlab}[1]{#1}
\providecommand{\url}[1]{\texttt{#1}}
\expandafter\ifx\csname urlstyle\endcsname\relax
  \providecommand{\doi}[1]{doi: #1}\else
  \providecommand{\doi}{doi: \begingroup \urlstyle{rm}\Url}\fi

\bibitem[Aytar et~al.(2017)Aytar, Castrejon, Vondrick, Pirsiavash, and
  Torralba]{aytar2017cross}
Y.~Aytar, L.~Castrejon, C.~Vondrick, H.~Pirsiavash, and A.~Torralba.
\newblock Cross-modal scene networks.
\newblock \emph{IEEE transactions on pattern analysis and machine
  intelligence}, 2017.

\bibitem[Bashkirova et~al.(2019)Bashkirova, Usman, and
  Saenko]{bashkirova2019adversarial}
D.~Bashkirova, B.~Usman, and K.~Saenko.
\newblock Adversarial self-defense for cycle-consistent gans.
\newblock In \emph{Advances in Neural Information Processing Systems}, pages
  635--645, 2019.

\bibitem[{Beaudry} and {Renner}(2011)]{beaudry2011dpi}
N.~J. {Beaudry} and R.~{Renner}.
\newblock {An intuitive proof of the data processing inequality}.
\newblock \emph{arXiv e-prints}, art. arXiv:1107.0740, July 2011.

\bibitem[Belghazi et~al.(2018)Belghazi, Baratin, Rajeswar, Ozair, Bengio,
  Courville, and Hjelm]{belghazi2018mine}
M.~I. Belghazi, A.~Baratin, S.~Rajeswar, S.~Ozair, Y.~Bengio, A.~Courville, and
  R.~D. Hjelm.
\newblock Mine: mutual information neural estimation.
\newblock \emph{arXiv preprint arXiv:1801.04062}, 2018.

\bibitem[Bousmalis et~al.(2017)Bousmalis, Silberman, Dohan, Erhan, and
  Krishnan]{bousmalis2017unsupervised}
K.~Bousmalis, N.~Silberman, D.~Dohan, D.~Erhan, and D.~Krishnan.
\newblock Unsupervised pixel-level domain adaptation with generative
  adversarial networks.
\newblock In \emph{Proceedings of the IEEE conference on computer vision and
  pattern recognition}, pages 3722--3731, 2017.

\bibitem[Choi et~al.(2018)Choi, Choi, Kim, Ha, Kim, and Choo]{choi2018stargan}
Y.~Choi, M.~Choi, M.~Kim, J.-W. Ha, S.~Kim, and J.~Choo.
\newblock Stargan: Unified generative adversarial networks for multi-domain
  image-to-image translation.
\newblock In \emph{Proceedings of the IEEE Conference on Computer Vision and
  Pattern Recognition}, pages 8789--8797, 2018.

\bibitem[Choi et~al.(2019)Choi, Uh, Yoo, and Ha]{choi2019stargan}
Y.~Choi, Y.~Uh, J.~Yoo, and J.-W. Ha.
\newblock Stargan v2: Diverse image synthesis for multiple domains.
\newblock \emph{arXiv preprint arXiv:1912.01865}, 2019.

\bibitem[Chu et~al.(2017)Chu, Zhmoginov, and Sandler]{chu2017cyclegan}
C.~Chu, A.~Zhmoginov, and M.~Sandler.
\newblock Cyclegan, a master of steganography.
\newblock \emph{arXiv preprint arXiv:1712.02950}, 2017.

\bibitem[Efros and Leung(1999)]{efros1999texture}
A.~A. Efros and T.~K. Leung.
\newblock Texture synthesis by non-parametric sampling.
\newblock In \emph{Proceedings of the seventh IEEE international conference on
  computer vision}, volume~2, pages 1033--1038. IEEE, 1999.

\bibitem[Heusel et~al.(2017)Heusel, Ramsauer, Unterthiner, Nessler, and
  Hochreiter]{heusel2017gans}
M.~Heusel, H.~Ramsauer, T.~Unterthiner, B.~Nessler, and S.~Hochreiter.
\newblock Gans trained by a two time-scale update rule converge to a local nash
  equilibrium.
\newblock In \emph{Advances in neural information processing systems}, pages
  6626--6637, 2017.

\bibitem[Huang et~al.(2018)Huang, Liu, Belongie, and
  Kautz]{huang2018multimodal}
X.~Huang, M.-Y. Liu, S.~Belongie, and J.~Kautz.
\newblock Multimodal unsupervised image-to-image translation.
\newblock In \emph{Proceedings of the European Conference on Computer Vision
  (ECCV)}, pages 172--189, 2018.

\bibitem[Isola et~al.(2017)Isola, Zhu, Zhou, and Efros]{pix2pix2016}
P.~Isola, J.-Y. Zhu, T.~Zhou, and A.~A. Efros.
\newblock Image-to-image translation with conditional adversarial networks.
\newblock \emph{Proceedings of the IEEE Conference on Computer Vision and
  Pattern Recognition}, 2017.

\bibitem[Kingma and Ba(2014)]{kingma2014adam}
D.~P. Kingma and J.~Ba.
\newblock Adam: A method for stochastic optimization, 2014.

\bibitem[Lee et~al.(2018)Lee, Tseng, Huang, Singh, and Yang]{lee2018diverse}
H.-Y. Lee, H.-Y. Tseng, J.-B. Huang, M.~Singh, and M.-H. Yang.
\newblock Diverse image-to-image translation via disentangled representations.
\newblock In \emph{Proceedings of the European Conference on Computer Vision
  (ECCV)}, pages 35--51, 2018.

\bibitem[Lee et~al.(2019)Lee, Tseng, Mao, Huang, Lu, Singh, and
  Yang]{lee2019drit++}
H.-Y. Lee, H.-Y. Tseng, Q.~Mao, J.-B. Huang, Y.-D. Lu, M.~Singh, and M.-H.
  Yang.
\newblock Drit++: Diverse image-to-image translation via disentangled
  representations.
\newblock \emph{arXiv preprint arXiv:1905.01270}, 2019.

\bibitem[Liu and Tuzel(2016)]{liu2016coupled}
M.-Y. Liu and O.~Tuzel.
\newblock Coupled generative adversarial networks.
\newblock In \emph{Advances in neural information processing systems}, pages
  469--477, 2016.

\bibitem[Liu et~al.(2017)Liu, Breuel, and Kautz]{liu2017unsupervised}
M.-Y. Liu, T.~Breuel, and J.~Kautz.
\newblock Unsupervised image-to-image translation networks.
\newblock In \emph{Advances in Neural Information Processing Systems}, pages
  700--708, 2017.

\bibitem[Liu et~al.(2015)Liu, Luo, Wang, and Tang]{liu2015deep}
Z.~Liu, P.~Luo, X.~Wang, and X.~Tang.
\newblock Deep learning face attributes in the wild.
\newblock In \emph{Proceedings of the IEEE international conference on computer
  vision}, pages 3730--3738, 2015.

\bibitem[Mao et~al.(2019)Mao, Lee, Tseng, Ma, and Yang]{mao2019mode}
Q.~Mao, H.-Y. Lee, H.-Y. Tseng, S.~Ma, and M.-H. Yang.
\newblock Mode seeking generative adversarial networks for diverse image
  synthesis.
\newblock In \emph{Proceedings of the IEEE Conference on Computer Vision and
  Pattern Recognition}, pages 1429--1437, 2019.

\bibitem[Oord et~al.(2018)Oord, Li, and Vinyals]{oord2018representation}
A.~v.~d. Oord, Y.~Li, and O.~Vinyals.
\newblock Representation learning with contrastive predictive coding.
\newblock \emph{arXiv preprint arXiv:1807.03748}, 2018.

\bibitem[Paszke et~al.(2017)Paszke, Gross, Chintala, Chanan, Yang, DeVito, Lin,
  Desmaison, Antiga, and Lerer]{paszke2017automatic}
A.~Paszke, S.~Gross, S.~Chintala, G.~Chanan, E.~Yang, Z.~DeVito, Z.~Lin,
  A.~Desmaison, L.~Antiga, and A.~Lerer.
\newblock Automatic differentiation in pytorch.
\newblock \emph{31st Conference on Neural Information Processing Systems (NIPS
  2017)}, 2017.

\bibitem[Poole et~al.(2019)Poole, Ozair, Oord, Alemi, and
  Tucker]{poole2019variational}
B.~Poole, S.~Ozair, A.~v.~d. Oord, A.~A. Alemi, and G.~Tucker.
\newblock On variational bounds of mutual information.
\newblock \emph{arXiv preprint arXiv:1905.06922}, 2019.

\bibitem[Rosales et~al.(2003)Rosales, Achan, and Frey]{rosales2003unsupervised}
R.~Rosales, K.~Achan, and B.~J. Frey.
\newblock Unsupervised image translation.
\newblock In \emph{iccv}, pages 472--478, 2003.

\bibitem[Sangkloy et~al.(2017)Sangkloy, Lu, Fang, Yu, and
  Hays]{sangkloy2017scribbler}
P.~Sangkloy, J.~Lu, C.~Fang, F.~Yu, and J.~Hays.
\newblock Scribbler: Controlling deep image synthesis with sketch and color.
\newblock In \emph{Proceedings of the IEEE Conference on Computer Vision and
  Pattern Recognition}, pages 5400--5409, 2017.

\bibitem[Shrivastava et~al.(2017)Shrivastava, Pfister, Tuzel, Susskind, Wang,
  and Webb]{shrivastava2017learning}
A.~Shrivastava, T.~Pfister, O.~Tuzel, J.~Susskind, W.~Wang, and R.~Webb.
\newblock Learning from simulated and unsupervised images through adversarial
  training.
\newblock In \emph{Proceedings of the IEEE conference on computer vision and
  pattern recognition}, pages 2107--2116, 2017.

\bibitem[Taigman et~al.(2016)Taigman, Polyak, and
  Wolf]{taigman2016unsupervised}
Y.~Taigman, A.~Polyak, and L.~Wolf.
\newblock Unsupervised cross-domain image generation.
\newblock \emph{arXiv preprint arXiv:1611.02200}, 2016.

\bibitem[Wang et~al.(2019)Wang, Jabri, and Efros]{wang2019learning}
X.~Wang, A.~Jabri, and A.~A. Efros.
\newblock Learning correspondence from the cycle-consistency of time.
\newblock In \emph{Proceedings of the IEEE Conference on Computer Vision and
  Pattern Recognition}, pages 2566--2576, 2019.

\bibitem[Yu et~al.(2019)Yu, Chen, Liu, Li, and Li]{yu2019multi}
X.~Yu, Y.~Chen, S.~Liu, T.~Li, and G.~Li.
\newblock Multi-mapping image-to-image translation via learning
  disentanglement.
\newblock In \emph{Advances in Neural Information Processing Systems}, pages
  2990--2999, 2019.

\bibitem[Zhai et~al.(2019)Zhai, Chen, Tung, He, Nawhal, and
  Mori]{zhai2019lifelong}
M.~Zhai, L.~Chen, F.~Tung, J.~He, M.~Nawhal, and G.~Mori.
\newblock Lifelong gan: Continual learning for conditional image generation.
\newblock In \emph{Proceedings of the IEEE International Conference on Computer
  Vision}, pages 2759--2768, 2019.

\bibitem[Zhang et~al.(2018)Zhang, Isola, Efros, Shechtman, and
  Wang]{zhang2018unreasonable}
R.~Zhang, P.~Isola, A.~A. Efros, E.~Shechtman, and O.~Wang.
\newblock The unreasonable effectiveness of deep features as a perceptual
  metric.
\newblock In \emph{Proceedings of the IEEE Conference on Computer Vision and
  Pattern Recognition}, pages 586--595, 2018.

\bibitem[Zhu et~al.(2017{\natexlab{a}})Zhu, Park, Isola, and
  Efros]{zhu2017unpaired}
J.-Y. Zhu, T.~Park, P.~Isola, and A.~A. Efros.
\newblock Unpaired image-to-image translation using cycle-consistent
  adversarial networks.
\newblock In \emph{Proceedings of the IEEE International Conference on Computer
  Vision}, pages 2223--2232, 2017{\natexlab{a}}.

\bibitem[Zhu et~al.(2017{\natexlab{b}})Zhu, Zhang, Pathak, Darrell, Efros,
  Wang, and Shechtman]{zhu2017multimodal}
J.-Y. Zhu, R.~Zhang, D.~Pathak, T.~Darrell, A.~A. Efros, O.~Wang, and
  E.~Shechtman.
\newblock Toward multimodal image-to-image translation.
\newblock In \emph{Advances in Neural Information Processing Systems},
  2017{\natexlab{b}}.

\bibitem[Zhu et~al.(2017{\natexlab{c}})Zhu, Zhang, Pathak, Darrell, Efros,
  Wang, and Shechtman]{zhu2017toward}
J.-Y. Zhu, R.~Zhang, D.~Pathak, T.~Darrell, A.~A. Efros, O.~Wang, and
  E.~Shechtman.
\newblock Toward multimodal image-to-image translation.
\newblock In \emph{Advances in Neural Information Processing Systems}, pages
  465--476, 2017{\natexlab{c}}.

\end{thebibliography}
}

\clearpage
\appendix

\section{Supplementary Overview}
This supplement provides the following additional information supporting the main paper.
\begin{list}{\textbullet}{}
  \item In \Cref{sec:extra_experiments}, we provide additional comparisons on the Photo$\rightarrow$Monet and Photo$\rightarrow$Portrait datasets. We also provide more generated images for Cat$\rightarrow$Dog, Summer$\rightarrow$Winter and CelebA attributes to supplement Section 4.2.
  \item \Cref{sec:extra_ablation} provides additional ablation studies to investigate the impact of the number of cycle translations, the effect of different mutual information bounds, and the impact of the margin value, to supplement Sections 3.1 and 3.3 of the main paper.
  \item \Cref{sec:tech_detail} provides technical details for the training pipeline described in Section 3.3.
\end{list}

\section{Additional Qualitative Results}
\label{sec:extra_experiments}

\xhdr{Photo$\rightarrow$Portrait and Photo$\rightarrow$Monet results.}
We perform additional qualitative comparisons on the Photo$\rightarrow$Portrait~\cite{zhu2017unpaired} and Photo$\rightarrow$Monet datasets~\cite{zhu2017unpaired}.
The comparisons to CycleGAN-based baselines follow the same settings as described in Section 4.1, where sampled images come from four different cycles.
On portrait datasets, we follow the original CycleGAN training procedure and introduce an additional identity loss in the first cycle.
See Section 5.2 and Figure 9 of the CycleGAN paper~\cite{zhu2017unpaired} for identity loss.

\begin{figure}
\centering
\includegraphics[width=0.95\textwidth]{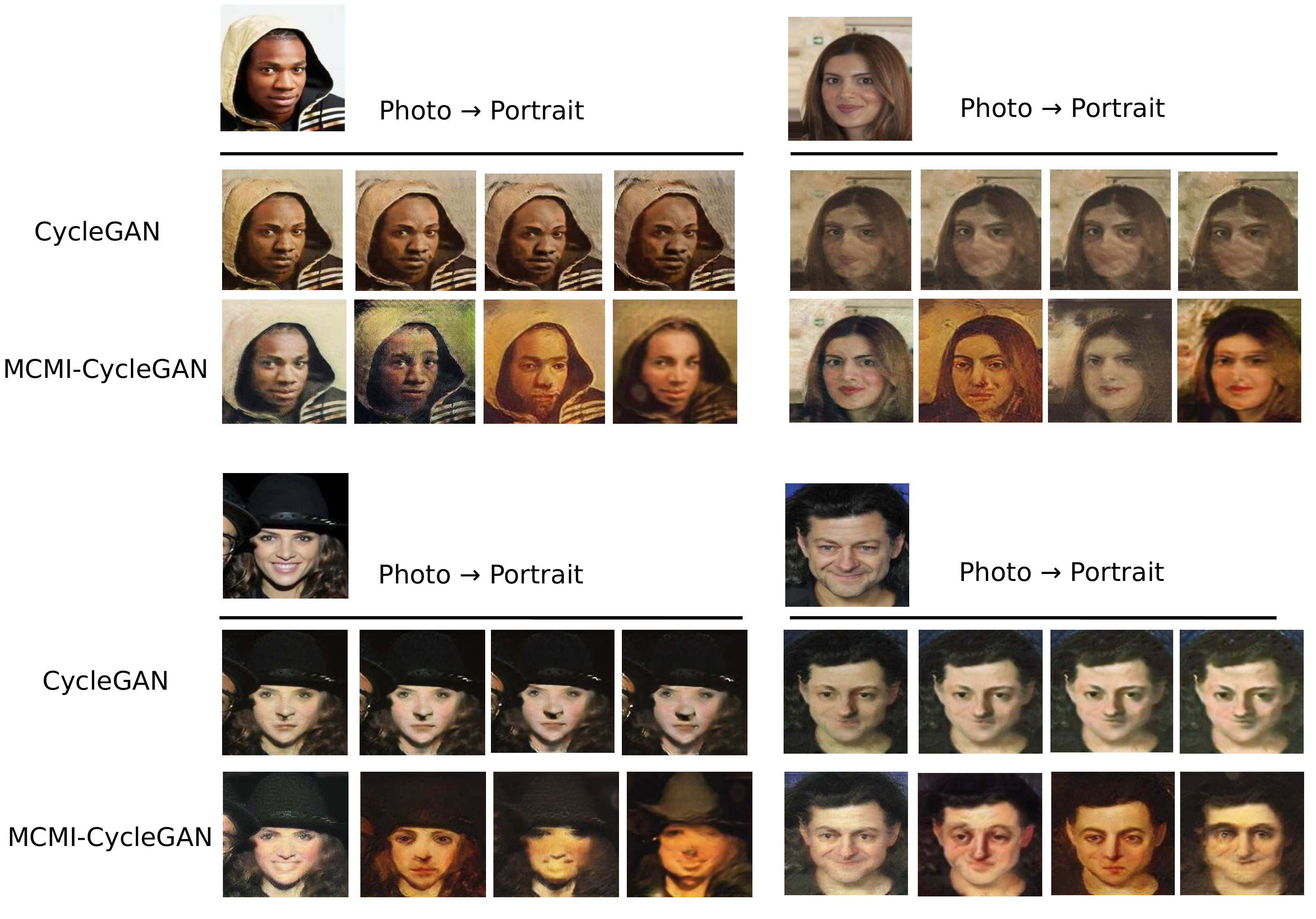}
\caption{\textbf{Photo$\rightarrow$Portrait results.}
Comparisons between images generated using the \mcmi approach and baseline modules. The top left image in each set is the input image. Then, each row shows output images corresponding to successive translation steps using
the specified model. 
}
\label{fig:portrait}
\end{figure}

Generated samples for Photo$\rightarrow$Portrait are in \Cref{fig:portrait}.
We observe that the single cycle CycleGAN baseline sometimes fails to generate portraits that have different art styles from the input photos.
Results from different cycles also look similar due to the fixed one-to-one mapping network structure.
However, when trained under the \mcmi framework, \mcmi-CycleGAN produces better translated images that look more like portrait paintings on the second cycle (second column for each input example).
We observe well-preserved facial structure, while there are changes to facial features such as perceived smoothness of the skin during multi-cycle translations.
Similarly to CycleGAN-based results on other datasets, the output of the first cycle supervised by self-reconstruction loss is of inferior quality as compared to the output of the second cycle.  

\begin{figure}
\centering
\includegraphics[width=0.95\textwidth]{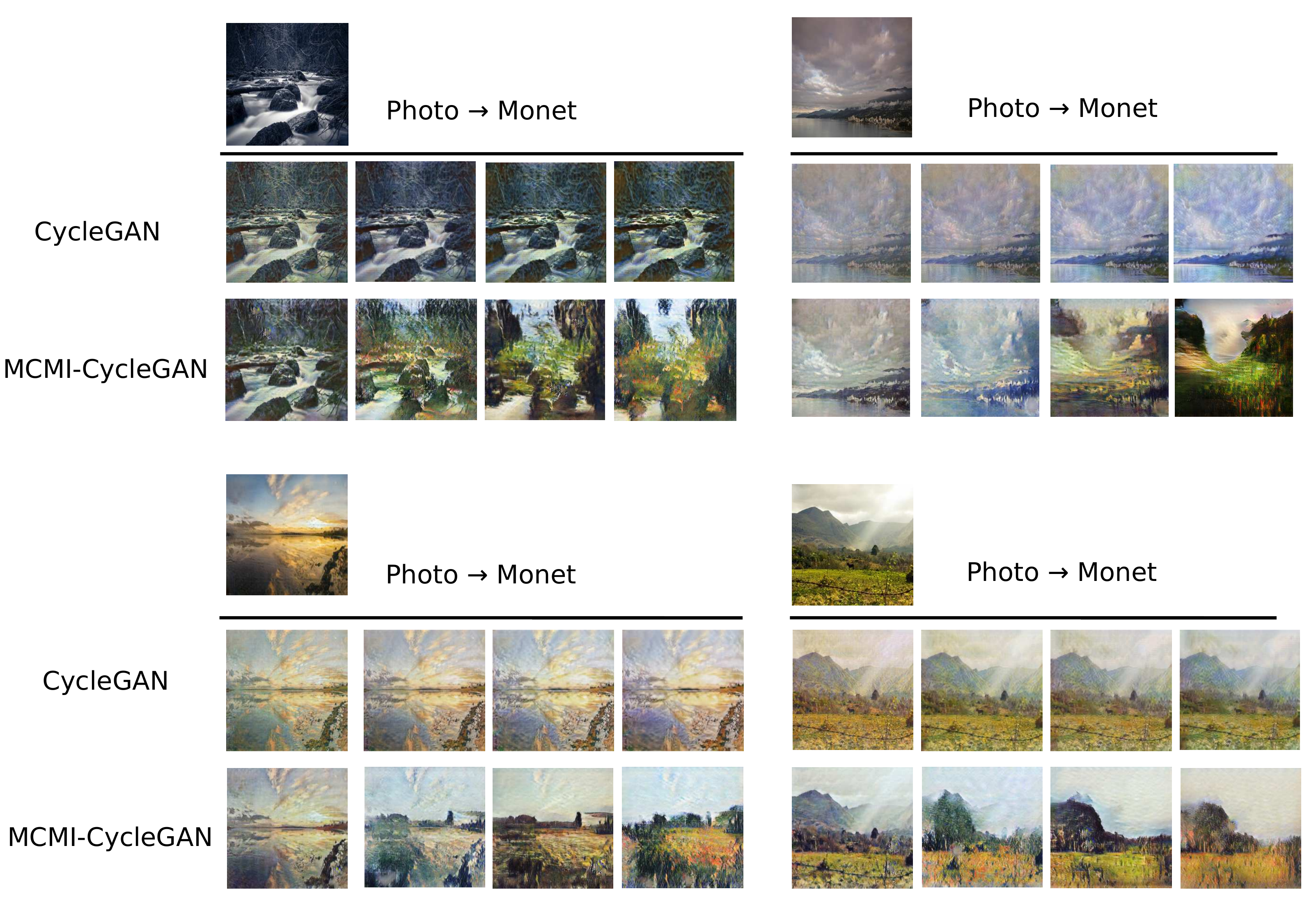}
\caption{\textbf{Photo$\rightarrow$Monet results.}
Comparisons between images generated using the \mcmi approach and baseline modules. The top left image in each set is the input image. Then, each row shows output images corresponding to successive translation steps using
the specified model. 
}
\label{fig:monet}
\end{figure}

\Cref{fig:monet} shows generated samples for Photo$\rightarrow$Monet.
It is difficult to reason about image quality for Monet paintings, but we do find that both \mcmi-CycleGAN and the baseline can preserve semantic structure of the input photo at the first and second cycles.
One interesting observation is that in later cycles, \mcmi-CycleGAN generates Monet images that look very different from the initial input.
For example, the sea surface in the bottom-left sample is slowly transformed into ground and the last few photos become a grass field rather than a lake.
This means that the model trained under the \mcmi framework has more diverse mappings and that later cross-domain translations only rely on the semantic information from the current cycle input, not the initial input where there is a lake in the scene.
This indicates more Markovian behavior for the model adapted using \mcmi.

\begin{figure}
\centering
\includegraphics[width=0.93\textwidth]{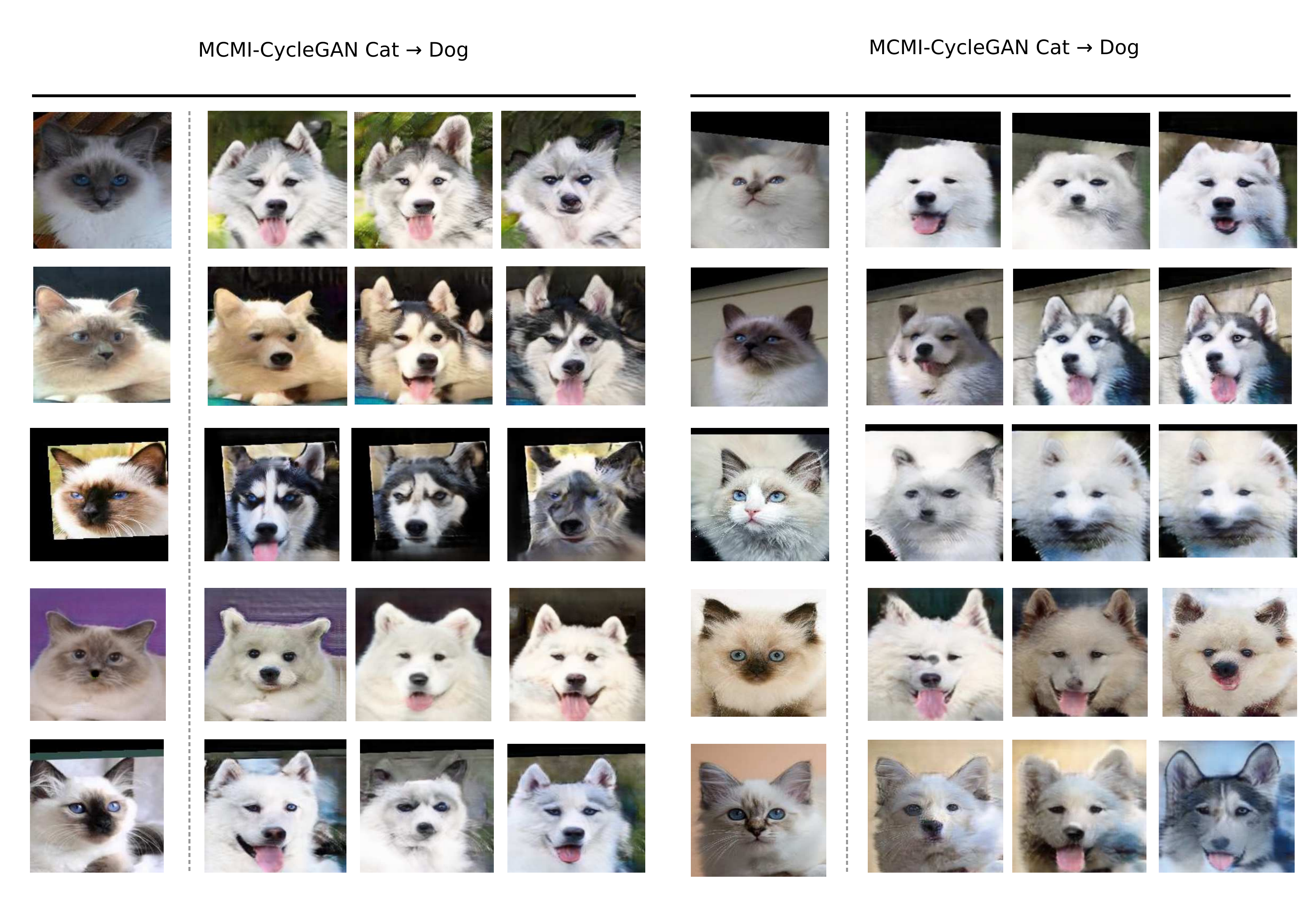}
\includegraphics[width=0.93\textwidth]{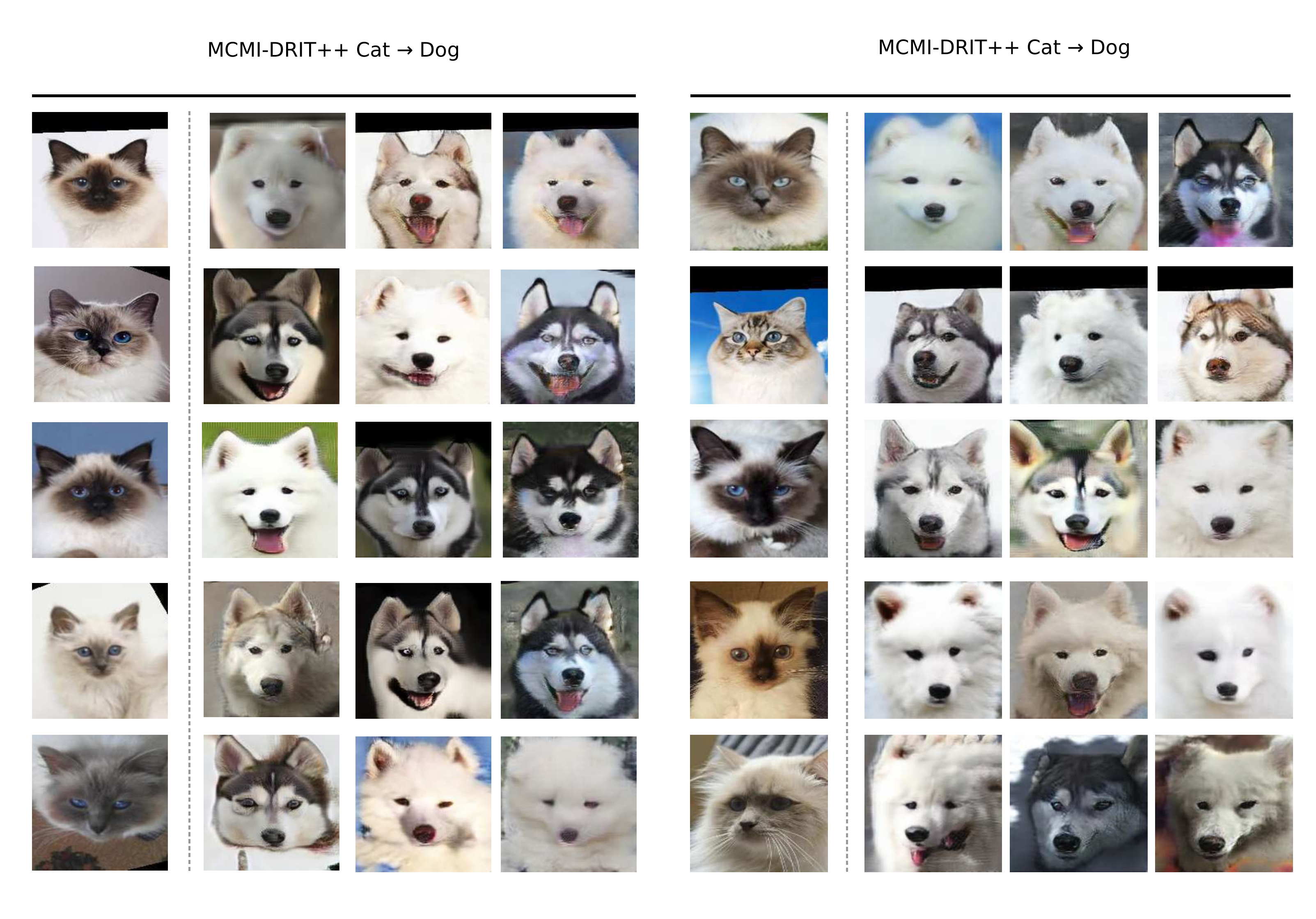}
\caption{\textbf{Additional Cat$\rightarrow$Dog results.}
The first image in each row is the input cat image.
Samples for \mcmi-CycleGAN are from successive output images after multiple translation steps (ignoring the first cycle).
Samples for \mcmi-DRIT++ come from different style codes at the first cycle.
}
\label{fig:qualitative_cat2dog_extra_no_compare}
\end{figure}

\xhdr{Additional Samples.}
We provide additional samples for the datasets introduced in Section 4.2.
\Cref{fig:qualitative_cat2dog_extra_no_compare} shows results for \mcmi-CycleGAN and \mcmi-DRIT++ on the Cat$\rightarrow$Dog dataset.
Following a similar setup as in Section 4.2, samples for \mcmi-CycleGAN are from successive output images at different cycles.
Section 4.2 demonstrates that first cycle \mcmi-CycleGAN results are inferior.
Thus, we ignore the first cycle and show results starting from the second cycle.  
Samples for the \mcmi-DRIT++ model come from different style codes at the first cycle.

\begin{figure}
\centering
\includegraphics[width=0.95\textwidth]{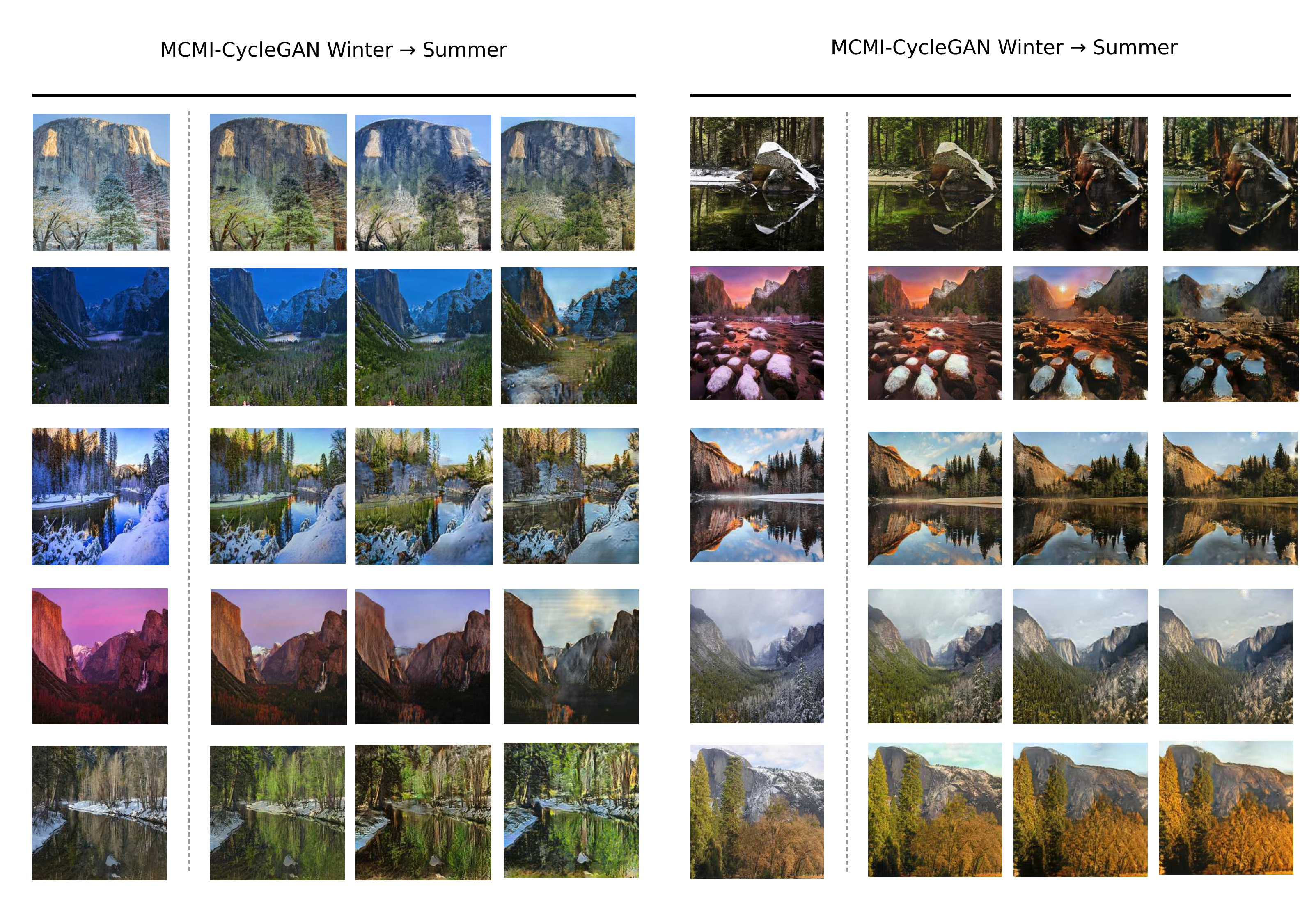}
\includegraphics[width=0.95\textwidth]{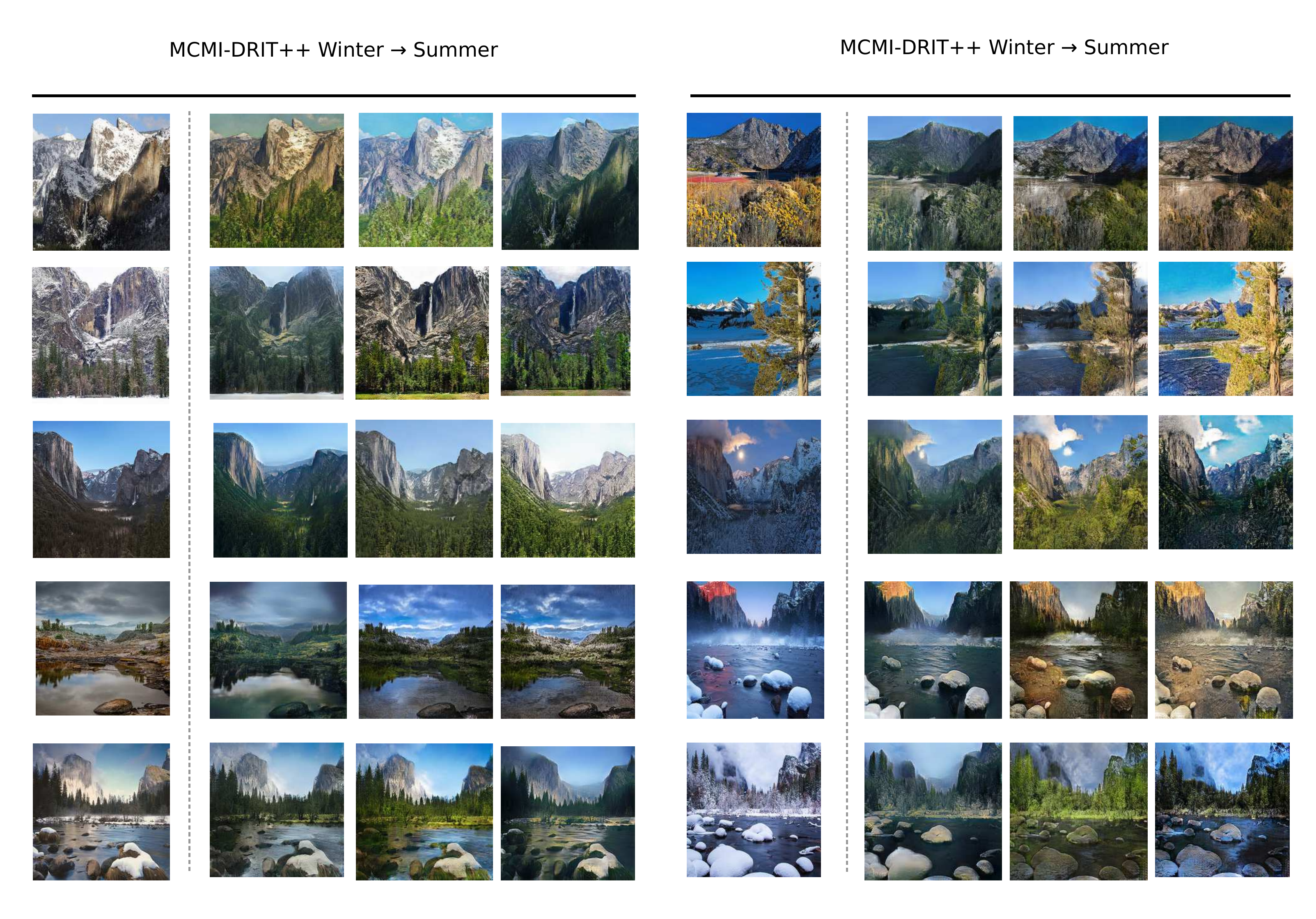}
\caption{\textbf{Additional Summer$\rightarrow$Winter Samples.}
First image at each row is the input winter image. Samples for \mcmi-CycleGAN are from successive output images after multiple translation steps (ignoring first cycle).
Samples for \mcmi-DRIT++ models come from different style codes at the first cycle.
}
\label{fig:qualitative_s2w_extra_no_compare}
\end{figure}

\Cref{fig:qualitative_s2w_extra_no_compare} provides additional results for \mcmi-CycleGAN and \mcmi-DRIT++ on the Summer$\rightarrow$Winter dataset.
We follow the standard setup for DRIT++~\cite{lee2019drit++} and use the simple concatenation model for fair comparison.

\begin{figure}
\centering
\includegraphics[width=1.0\textwidth]{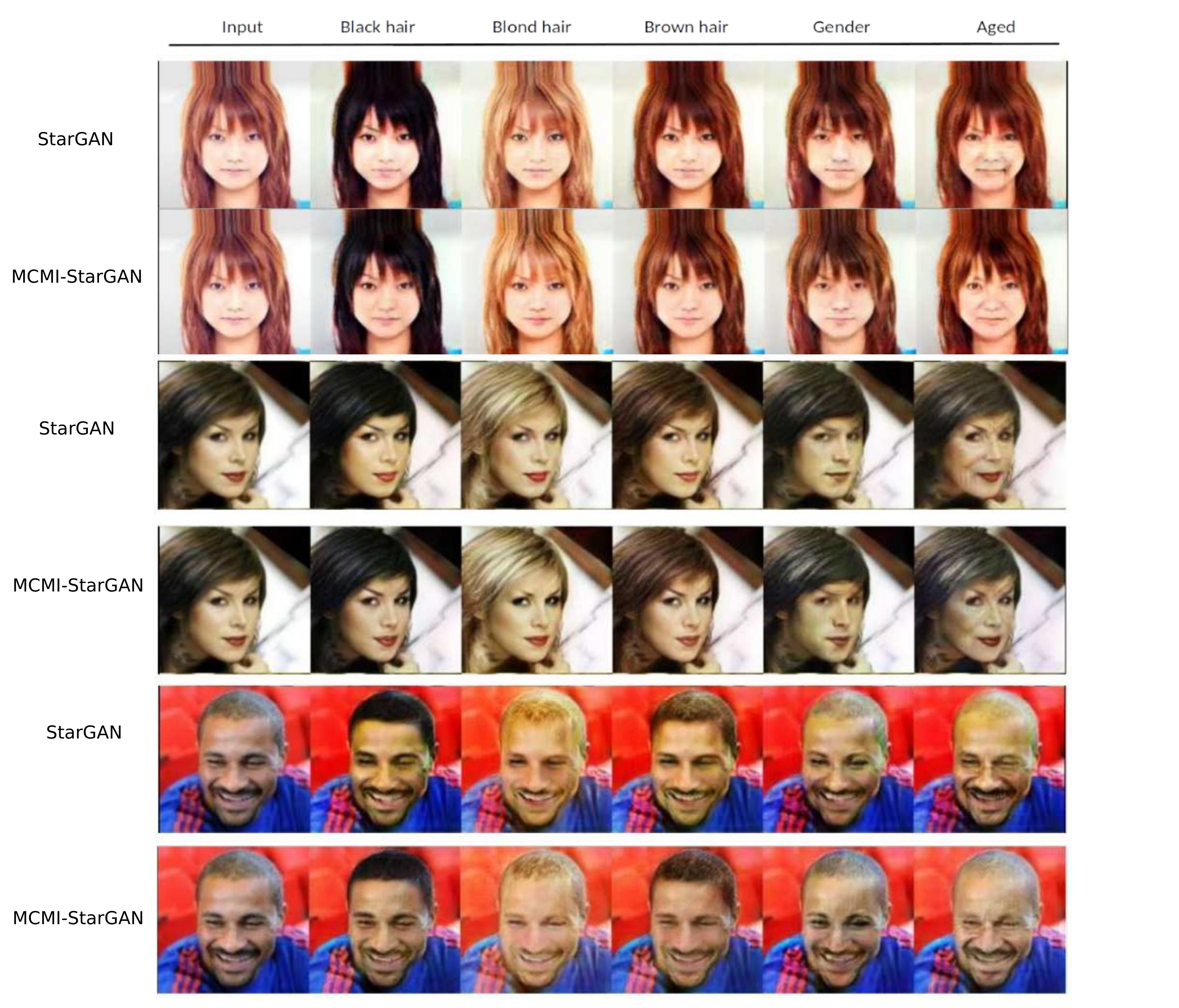}
\caption{
\textbf{Additional multi-domain I2I translation results on the CelebA~\cite{liu2015deep} dataset.}
Each pair of rows compares the baseline StarGAN model with an \mcmi-adapted StarGAN model.
}
\label{fig:qualitative_stargan_extra}
\end{figure}

\Cref{fig:qualitative_stargan_extra} shows additional results for \mcmi-StarGAN trained on the CelebA dataset~\cite{liu2015deep}.
We observe that our \mcmi framework produces more realistic wrinkles on aged pictures and is better at modeling the hair.

\begin{figure}
\includegraphics[width=1.0\textwidth]{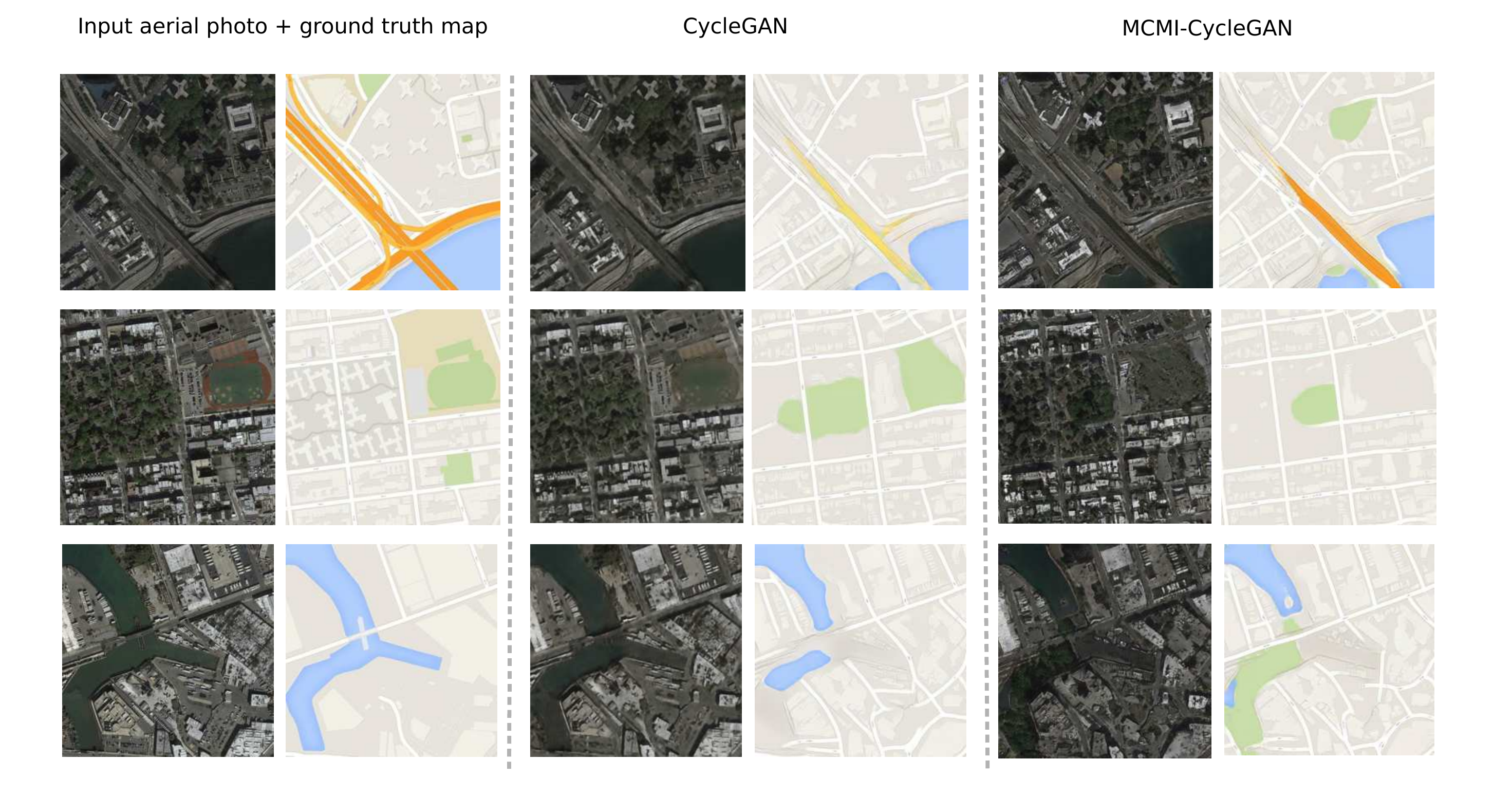}
\caption{\textbf{Additional Aerial photo to map results}. \mcmi-CycleGAN better preserves the semantic information from generated maps to reconstructed photo, exhibiting more Markovian behavior.
}
\label{fig:qualitative_maps_extra}
\end{figure}

\Cref{fig:qualitative_maps_extra} provides more samples of CycleGAN-based models on the Google Aerial Photo to Maps dataset~\cite{zhu2017unpaired}.
We observe that the translated map and reconstructed aerial photo from \mcmi-CycleGAN have more Markovian behavior.
For instance in the last row, the lower-left part of the translated map contains trees instead of lake, and the reconstructed aerial photo then reflects this difference.
On the other hand, the standard CycleGAN model reconstructs the lake even though the translated map contains no lake in the bottom-left region.

\begin{figure}
\includegraphics[width=1.0\textwidth]{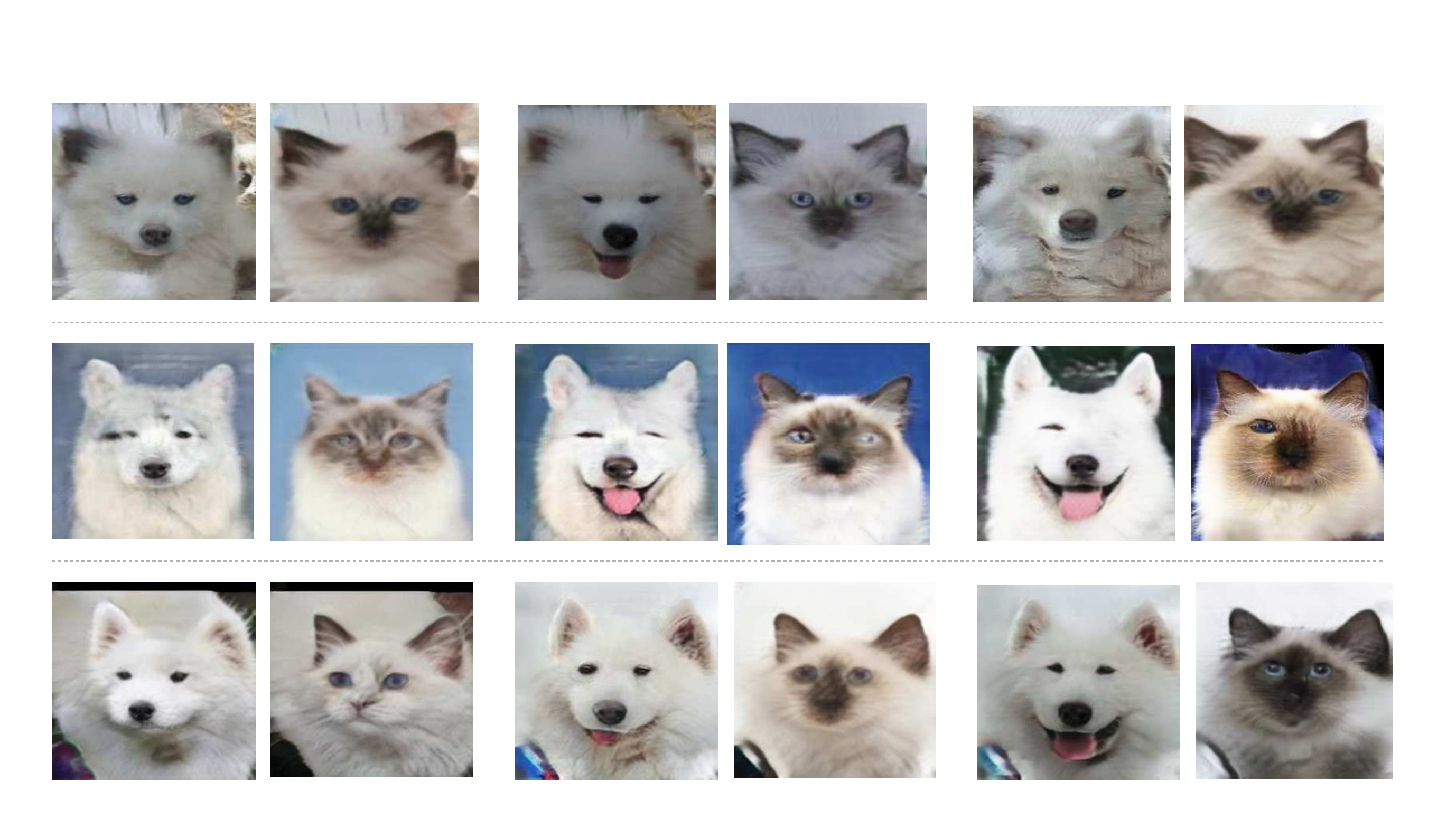}
\caption{\textbf{Markovian behavior on Cat$\rightarrow$Dog dataset}. Each row represents one multi-cycle translation with corresponding cat and dog images at each cycle.}
\label{fig:qualitative_m_chain}
\end{figure}

To further demonstrate Markovian behavior from our \mcmi-adapted models, we also show the corresponding input cat image and generated dog image at each cycle on the Cat$\rightarrow$Dog dataset. \Cref{fig:qualitative_m_chain} illustrates three different multi-cycle translations from \mcmi-CycleGAN.
We can see that when the associated cat image changes, the translated dog image is also different at each cycle.
For instance, in the second row, when one of the eyes begins to disappear, subsequent results also have a disappearing eye.

\section{Additional Ablation Studies}
\label{sec:extra_ablation}

\begin{figure*}
\includegraphics[width=0.5\textwidth]{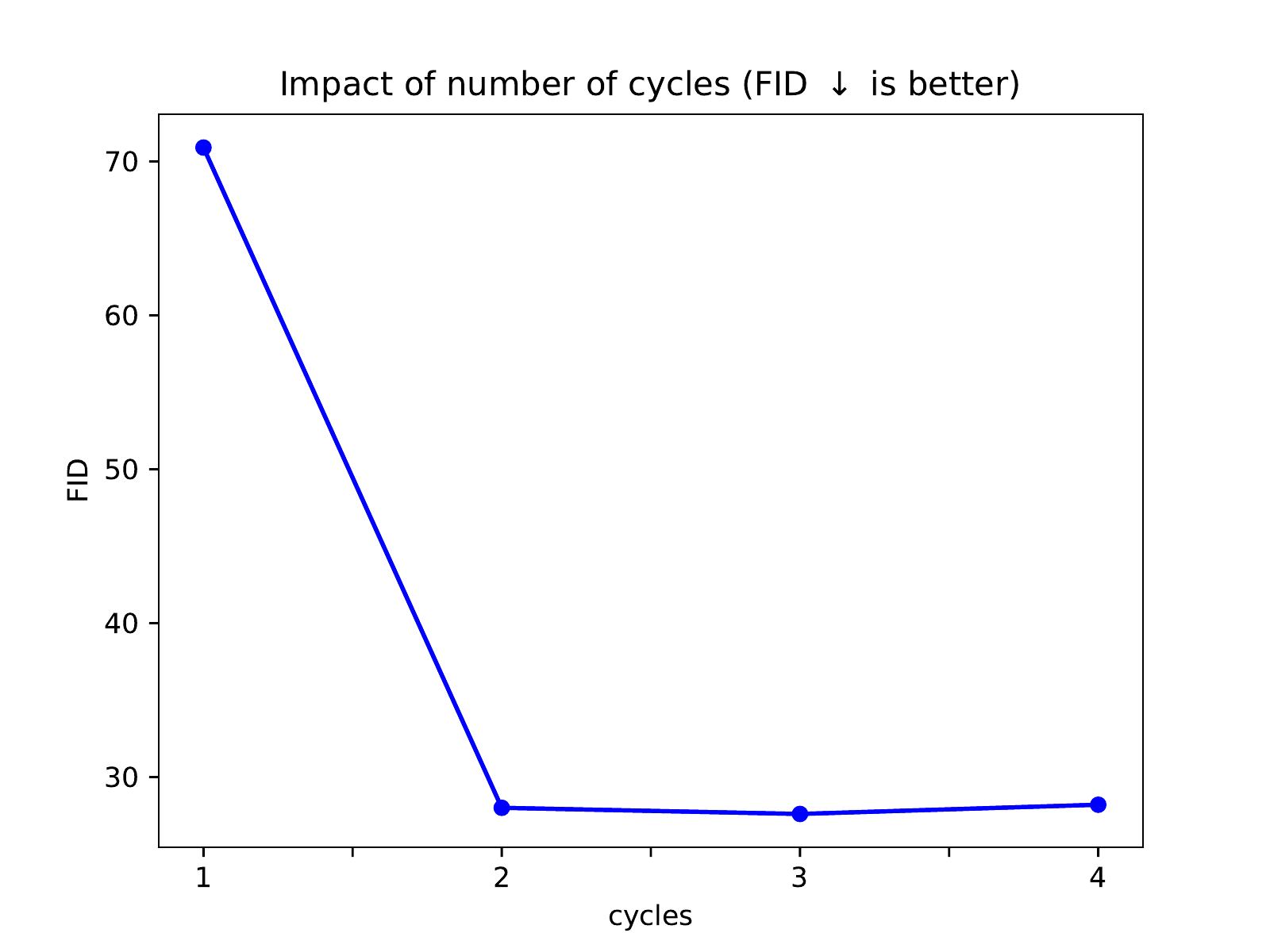}
\includegraphics[width=0.5\textwidth]{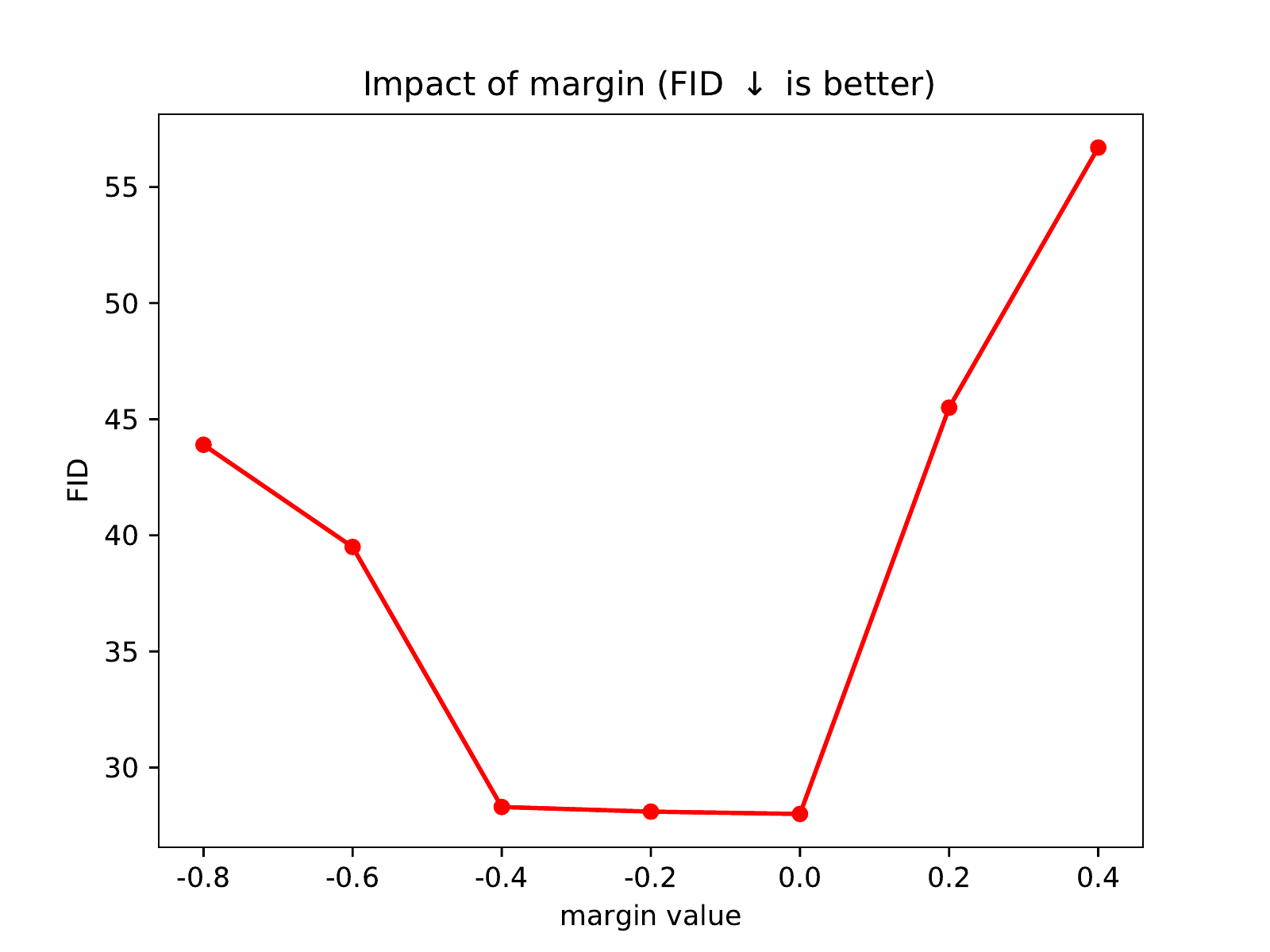}
\caption{
\textbf{Effect of number of cycle translations (left) and the MI constraint margin (right).}
FID scores are reported for the \mcmi-CycleGAN model trained on the Cat$\rightarrow$Dog dataset.
On the left, we observe that performance gain is not significant after two cycles.
On the right, we find that a small margin between $-0.4$ and $0.0$ offers the best performance.
}
\label{fig:ablats_mi}
\end{figure*}

\xhdr{Impact of the number of cycle translations.}
We enforce our proposed mutual information constraints on different numbers of cycle translations and study the effect on model performance.
We conduct this ablation with the \mcmi-CycleGAN model on the Cat$\rightarrow$Dog dataset.
\Cref{fig:ablats_mi} (left) plots the FID score across different numbers of translation cycles.
Note that the non-increasing mutual information constraints are enforced.
We observe that the performance gain is not significant after two cycles of translations.
However, GPU memory use increases for each additional translation step.
This supports that our choice of two cycle \mcmi in Section 4.5 offers a good trade-off between performance and memory.

\xhdr{Effect of the MI margin.}
Equation 4 from the main paper strictly enforces the non-increasing MI constraints with upper and lower bounds.
However, this constraint might be too hard.
It is possible to add a margin to $L_{\text{\tiny MI}}$.
A negative margin means that we relax the constraint and a positive margin means that MI must decrease by a certain value.
For this ablation study, we use different values of the margin and evaluate their effect on the model performance.
Similar to before, we use CycleGAN as our backbone I2I module and train the model on the Cat$\rightarrow$Dog dataset.
\Cref{fig:ablats_mi} (right) shows the FID score plotted against different margin values.
We see that a small margin value between $-0.4$ and $0.0$ results in the best performance.

\xhdr{Effect of different MI bounds.}
In \Cref{tab:ablation_constraint}, we report model performance on the Cat$\rightarrow$Dog dataset with different versions of the MI bounds used in the training loss.
We first compare to only increasing the MI lower bound without any MI constraints (CycleGAN + MI).
We find that this leads to a decrease in diversity in the generated images (indicated by lower LPIPS) as well as less realistic generation (shown by higher FID1 and FID2 scores).
We then compare \mcmi-CycleGAN with upper and lower bounds to using both lower bounds in the MI constraints (lower bounds MI), enforcing the exact opposite MI constraints (non-decreasing MI), and enforcing MI to be the same across cycles (non-changing MI).
Similarly to before, all other versions result in decrease in LPIPS and worse FID scores for the translated images obtained after the first and second cycles.
We only observe minor improvement in FID1 for non-changing MI.
However, FID2 and LPIPS are much worse than before.

\begin{table*}[t]
\renewcommand{\arraystretch}{1.3}
\centering
\caption{\textbf{CycleGAN-based models with different implementation of MI loss functions}. Arrows indicate whether lower ($\downarrow$) or higher ($\uparrow$) value is better. Standard deviations of test set numbers indicated for FID. All standard deviations for LPIPS values are $\leq 0.003$.}
\vspace{2pt}
\begin{tabular}{@{} l @{\hspace{20pt}} ccc @{}}
\toprule
 & \multicolumn{3}{c}{Cat$\rightarrow$Dog} \\
 \cmidrule(l){2-4} 
Model & FID1$\downarrow$  & FID2$\downarrow$ & LPIPS$\uparrow$~ \\\midrule
CycleGAN + MI & $69.6\pm0.32$ &$35.0\pm0.33$ & $0.10$ \\ 
\mcmi-CycleGAN & $62.9\pm0.34$ & $\mathbf{28.0\pm0.34}$ & $\mathbf{0.22}$ \\ 
\mcmi-CycleGAN (lower bounds MI)& $65.1\pm0.28$ &$32.3\pm0.26$ & $0.20$ \\ 
\mcmi-CycleGAN (non-decreasing MI)& $88.9\pm0.44$ &$59.6\pm0.39$ & $0.19$ \\ 
\mcmi-CycleGAN (non-changing MI)& $\mathbf{58.7\pm0.30}$ & $36.0\pm0.31$ & $0.11$ \\

\bottomrule
\end{tabular}
\label{tab:ablation_constraint}
\end{table*}

\section{Technical Details}
\label{sec:tech_detail}

As described in Section 3.3, the mutual information evaluation module consists of four convolutional layers with instance batch normalization.
The number of channels per layer is $64\rightarrow128\rightarrow128\rightarrow128$.
We use a leaky ReLU as our activation function and max-pooling in earlier layers.
The mutual information estimation network directly takes an entire image as input.
During training, we separately optimize the I2I translation module and MI estimation module at two different steps.
In the first step, we compute the \mcmi loss $L_{\text{\tiny MI}}$ and only optimize the I2I model with this gradient.
In the second step, we compute $I_\text{lower}$ and optimize the MI estimation model so that the MI lower bound is increased and the critic is closer to the optimal.

Note that the gradient from the MI estimation step does not flow back to the I2I translation model as we optimize the two modules separately.
Only the \mcmi loss $L_{\text{\tiny MI}}$ optimizes the I2I model.
InfoNCE~\cite{oord2018representation} can not provide an accurate MI estimation with a small batch size.
Thus, during training we also provide additional anchor images as input to the MI estimation network.
We use a batch size of $8$ in all our experiments.
For training \mcmi-StarGAN on CelebA, we also found that the additional image adversarial loss leads to convergence issues, and do not use it.

We use the Adam~\cite{kingma2014adam} optimizer to train our models.
The learning rate and decay schedule are kept the same as originally used for each baseline module.
The learning rate is set to $1e^{-4}$ for DRIT++, MUNIT, and StarGAN-based models.
The learning rate is $2e^{-4}$ for CycleGAN-based models.
We train DRIT++ models for $1.2$k epochs, MUNIT models for $50$k iterations, StarGAN models for $20$k iterations, and CycleGAN models for $200$ epochs.
The batch size for the I2I model is $1$ for CycleGAN-based and MUNIT-based models and $2$ for DRIT++ models.
In Equation 5 from the main paper, we set $\alpha$ to $0.5$ and $\beta$ to $1.0$.

For evaluation, we follow a similar approach as in previous works.
For DRIT++ and MUNIT-based models, we first extract the content feature from the input image.
To translate an image, we randomly sample the attribute (or style) feature and combine it with the encoded content feature before passing them to the image generator.
For computing FID2, output samples from the previous translation step are used as input for the next cycle.
For StarGAN and CycleGAN-based models that do not support feature disentanglement, we directly pass the output sample from the previous translation step to the image generator.

\end{document}